%% file: main.tex
\pdfoutput=1
\documentclass{article}
\PassOptionsToPackage{numbers}{natbib}
\usepackage[final]{neurips_2024}

\usepackage[utf8]{inputenc} 
\usepackage[T1]{fontenc}    
\usepackage{hyperref}
\hypersetup{pagebackref,breaklinks,colorlinks,citecolor=blue}

\usepackage{url}            
\usepackage{booktabs}       
\usepackage{amsfonts}       
\usepackage{nicefrac}       
\usepackage{microtype}      
\usepackage{xcolor}         
\usepackage{subcaption}

\DeclareUnicodeCharacter{03C0}{\ensuremath{\pi}}

\usepackage{mathtools}
\usepackage[pdftex]{graphicx}
\usepackage{makecell}
\usepackage{amsmath}
\usepackage{dsfont}
\usepackage{bbold}
\usepackage{rotating}
\usepackage{multirow}
\usepackage{colortbl} 
\usepackage{tabularx}
\usepackage{pifont}
\usepackage{wrapfig}
\usepackage{enumitem}

\input{tex/abbrev}

\title{\vm{} and \vam{}: Autonomous Driving through Video Generative Modeling}

\author{%
  Florent Bartoccioni$^\dagger$ \\
  \And
  Elias Ramzi$^\dagger$
  \And
  Victor Besnier
  \And
  Shashanka Venkataramanan
  \And
  Tuan-Hung Vu
  \And
  Yihong Xu
  \And
  Loick Chambon$^1$
  \And
  Spyros Gidaris
  \And
  Serkan Odabas
  \And
  David Hurych
  \And
  Renaud Marlet$^2$
  \And
  Alexandre Boulch
  \And
  Mickael Chen$^\star$
  \And
  \'Eloi Zablocki
  \And
  Andrei Bursuc
  \And
  Eduardo Valle
  \And
  Matthieu Cord$^1$
}

\usepackage{xspace}
\newcommand{\vm}{VaViM\xspace}
\newcommand{\vam}{VaVAM\xspace}

\begin{document}

\maketitle

\vspace{-2em}
\begin{center}
    valeo.ai, Paris, France \\[1em]
    Project page: \url{https://valeoai.github.io/vavim-vavam/}
\end{center}

\newcommand\blfootnote[1]{%
  \begingroup
  \renewcommand\thefootnote{}\footnote{#1}%
  \addtocounter{footnote}{-1}%
  \endgroup
}

\blfootnote{$^\dagger$ Corresponding authors; \texttt{\{florent.bartoccioni, elias.ramzi\}@valeo.com}}
\blfootnote{$^\star$ Work done while at valeo.ai, now at H company.}
\blfootnote{$^1$ valeo.ai \& Sorbonne Université, Paris, France}
\blfootnote{$^2$ valeo.ai \& ENPC, Paris, France}

\input{tex/abstract}

\input{tex/intro}
\input{tex/related}

\input{tex/method}
\input{tex/data}
\input{tex/eval}

\input{tex/conclusion}
{\small
\bibliographystyle{plain}
\bibliography{biblio}
}

\end{document}

%% file: tex/abbrev.tex
\definecolor{th}{RGB}{220, 220, 220} 
\definecolor{lightblue}{RGB}{173, 216, 230}
\definecolor{nicegreen}{RGB}{102, 204, 102}
\definecolor{darkgreen}{RGB}{0, 128, 0}

\newcolumntype{Y}{>{\centering\arraybackslash}X}
\newcolumntype{T}{>{\centering\small\arraybackslash}X}
\newcolumntype{L}{>{\centering\small\arraybackslash}l}

\newcommand{\Th}[1]{\textsc{#1}}

\definecolor{valeocolor}{RGB}{130, 230, 0}
\colorlet{valeocell}{valeocolor!20}

\definecolor{eloiblue}{RGB}{0, 0, 165}

\def\ie{\textit{i.e.}\xspace}
\def\eg{\textit{e.g.}\xspace}
\def\etc{\textit{etc.}\xspace}

%% file: tex/abstract.tex
\begin{abstract}

We explore the potential of large-scale generative video models for autonomous driving, introducing an open-source auto-regressive video model (\vm) and its companion video-action model (\vam) to investigate how video pre-training transfers to real-world driving. \vm is a simple auto-regressive video model that predicts frames using spatio-temporal token sequences. We show that it captures the semantics and dynamics of driving scenes.
\vam{}, the video-action model, leverages the learned representations of \vm to generate driving trajectories through imitation learning. Together, the models form a complete perception-to-action pipeline.
We evaluate our models in open- and closed-loop driving scenarios, revealing that video-based pre-training holds promise for autonomous driving.
Key insights include the semantic richness of the learned representations, the benefits of scaling for video synthesis, and the complex relationship between model size, data, and safety metrics in closed-loop evaluations.
We release code and model weights at \href{https://github.com/valeoai/VideoActionModel}{\nolinkurl{github.com/valeoai/VideoActionModel}}.
\end{abstract}

%% file: tex/intro.tex

\section{Introduction}
\label{sec:intro}

Large-scale generative models have shattered the status quo of video generation with photorealistic, temporally coherent, high-fidelity videos synthesized from textual prompts.
While generalist models such as Sora, Veo-2 \citep{veoteam2024veo2} and VideoJAM \citep{chefer2025videojam} demonstrate those capabilities at large, specialist models such as GAIA-1 \citep{hu2023gaia1} and VISTA \citep{gao2024vista} showcase impressive performance in predicting future frames of driving videos.
Generating plausible future frames suggests that those models capture meaningful representations of the world, but the exact nature of such representations, as well as any practical utility they might have for actual driving, remain open questions. To what extent do those representations encode driving-relevant features, such as scene dynamics, geometry, and semantics? How far do they apply to actual autonomous systems, enhancing downstream tasks, such as motion planning?

To answer those questions, we introduce an open-source, large-scale autoregressive video model (\vm) and its companion video-action model (\vam).
At the core of our approach, \vm learns to predict future frames by modeling the joint distribution of spatio-temporal token sequences, capturing the underlying dynamics of driving scenes into dense representations. We use an image tokenizer to compress visual information into discrete tokens, providing a compact representation of each video frame.
To bridge the gap between video understanding and action generation, we train an action expert module on VaViM's learned video representation, forming our complete VaVAM system. We train that module with imitation learning to generate future driving trajectories guided by high-level goals and temporal context extracted from \vm{}. That architecture forms a complete perception-to-action pipeline, enabling effective motion planning and decision-making in autonomous vehicles.

Our work is the first large-scale study to explore how video generative pre-training transfers to driving capabilities. We evaluate our approach using both open- and closed-loop driving scenarios. Our findings suggest that video-based pre-training holds great promise for autonomous driving, with the following insights:
\begin{enumerate}
    \item The learned representations from video pre-training contain rich and meaningful semantic information.
    \item Larger models generally improve video synthesis quality, reinforcing the benefits of scale for generative modeling. However, larger video models perform worse than smaller ones on semantic segmentation tasks, suggesting that better generative quality does not directly translate to better semantic understanding.
    \item Scaling up the model improves performances in open-loop evaluations, and increasing training data yields further improvements. Nonetheless, scaling model size or data does not consistently improve safety metrics in closed-loop evaluations. This reveals a fundamental conflict between trajectory-following and adaptive decision-making.
\end{enumerate}

Our key contributions are as follows:
\begin{itemize}
    \item We provide data mix, scaling laws, training recipes, and a detailed reproducible protocol for training an autoregressive Video Model (\vm{}) on large-scale 1800+ hours diverse driving data.
    \item We present a procedure to adapt a video model into a video-action model (\vam{}) using imitation learning for end-to-end driving from camera input.
    \item We propose new evaluations for the learned \vm{} representations to assess their semantic content. Additionally, we benchmark \vam{} in both open- and closed-loop driving scenarios, emphasizing safety-critical situations. \vam{} achieves state-of-the-art performance in frontal driving scenarios on NeuroNCAP \citep{ljungbergh2024neuroncap}.
\end{itemize}

\autoref{tab:model_overview} reports all \vm{} and \vam{} models produced in this research, coming in different sizes. We release both the source code and the weights for those models.

\begin{table*}[h]
\caption{
Overview of the released models, covering different model sizes (up to 1.2B), trained on increasing amounts of data, and two model types (video generation and action learning).
}
\label{tab:model_overview}
\centering
\begin{tabular}{l l c c c}
\toprule
\Th{Model} & \Th{Parameters (M)} & \Th{OpenDV} \citep{yang2024opendv} & \Th{nuPlan} \citep{caesar2021nuplan} & \Th{nuScenes} \citep{caesar2020nuscenes} \\
& & \textit{1700+ hours} & \textit{94 hours} & \textit{5.5 hours} \\
\midrule
\multicolumn{5}{c}{\cellcolor{valeocell} \texttt{Pre-trained video models}} \\
VaViM-S & 185 & \ding{51} & & \\
VaViM-B & 318 & \ding{51} & & \\
VaViM-L & 1,200 & \ding{51} & & \\
\midrule
\multicolumn{5}{c}{\cellcolor{valeocell} \texttt{Fine-tuned video models}} \\
VaViM-S & 185 & \ding{51} & \ding{51} & \ding{51} \\
VaViM-B & 318 & \ding{51} & \ding{51} & \ding{51} \\
VaViM-L & 1,200 & \ding{51} & \ding{51} & \ding{51} \\
\midrule
\multicolumn{5}{c}{\cellcolor{valeocell} \texttt{Video-action models trained with imitation learning}} \\
VaVAM-S & 185 + 21 &  & \ding{51} & \ding{51} \\
VaVAM-B & 318 + 38 &  & \ding{51} & \ding{51} \\
VaVAM-L & 1,200 + 150 &  & \ding{51} & \ding{51} \\
\bottomrule
\end{tabular}
\end{table*}

%% file: tex/related.tex
\section{Related work}
\label{sec:related}

\subsection{Video generative model}

Today's sophisticated architectures for video generation have significantly evolved since the days of Generative Adversarial Networks (GANs)~\cite{goodfellow2020gan, saito2017tgan, tulyakov2018mocogan}. Beyond basic generation capabilities, recent advances have focused on improving visual quality and enabling precise control over generated content. That dual evolution is visible in key developments in video data representation and video generation aligned to conditioning signals.

\paragraph{Continuous Representations:} These methods work with real-valued embeddings in a continuous latent space, often operating on a pre-trained Variational Autoencoder (VAE)~\cite{kingma2013vae,higgins2017betavae} to compress the image or video signal spatially and reduce computation.  
Most modern approaches in this category build on either Diffusion or Flow Matching models. Diffusion Models (DM)~\cite{ho2020ddpm,rombach2022ldm} learns to model the data distribution through a denoising process, effectively capturing the data distribution of images. Methods such as Video Diffusion Model (VDM)~\cite{ho2022vdm}, Make-A-Video~\cite{singer2023makeavideo}, or Align Your Latents~\cite{blattmann2023align} extends image diffusion model to video generation 
Flow Matching~\cite{lipman2023flow} is similar to DM in learning continuous probability flows but contrasts with the latter by directly modeling the noise-to-image vector field.
Notable works such as MovieGen~\cite{polyak2024movie} or pyramid flow matching~\cite{jin2024pyramidal} adapt Flow Matching to generate videos efficiently.

\paragraph{Discrete Representations:} These approaches typically map video data into sequences of discrete tokens using vector quantization techniques (e.g., VQ-VAE~\cite{van2017vqvae}, FSQ~\cite{menter2024fsq}, LFQ~\cite{yu2024magvit2} etc.). Inspired by language models, auto-regressive methods generate tokens sequentially, conditioning each token on the previous ones. Building upon the success of auto-regressive image generation~\cite{chen2020igpt,ramesh2021dalle,esser2020taming}, notable extensions to video generation include VideoGPT~\cite{yan2021videogpt} and HARP~\cite{Seo2022HARP} where the generation becomes spatio-temporal. 
In contrast to purely sequential prediction, Masked Image Modeling~\cite{chang2022maskgit} uses a mask-and-predict strategy that iteratively predicts/reconstructs the missing tokens, inspired by BERT-like~\cite{devlin2019BERT} training schemes. 
Models such as MagVIT~\cite{yu2023magvit} and its improvement MagVIT-v2~\cite{yu2024magvit2} have effectively extended such framework to video generation, while Phenaki~\cite{villegas2023phenaki} has used it to enable long video generation from text prompts.

\paragraph{Action-Controlled Video Generation:} 

Integrating action-based control marks a significant advance in video generation, promising to move beyond passive generation to planning and decision-making.
That direction is exemplified by modern `neural game engines'~\cite{Bamford2020neuralgameengine}, such as Genie~\cite{bruce2024genie}, DIAMOND~\cite{alonso2024DIAMOND}, or GameNGen~\cite{valevski2024GameNGen}, which generate, from input actions, the evolution of complex, dynamic, and tridimensional game environments. 
In the automotive context, state-of-the-art approaches build from driving videos towards `neural world simulators' or so-called `world-models' by leveraging pre-trained diffusion networks, as VISTA \citep{gao2024vista}, GEM \citep{hassan2024gem}, and  InfinityDrive \citep{guo2024infinitydrive} do, or by exploiting discrete-based auto-regressive models, as GAIA-1 \cite{hu2023gaia1} and DrivingWorld \citep{hu2024drivingworld} do. 
However, a significant gap remains between generating realistic videos and learning representations suitable for robust decision-making. That limitation appears in current evaluation methods, which primarily use metrics that assess the visual quality~\citep{Unterthiner2019FVD,zhang2018lips} but provide limited insight into the performance of end tasks.

\subsection{Scalable Vision-based Action Learning}

Traditional approaches to vision-based action learning have relied heavily on expensive human annotations such as semantic segmentation masks, bounding boxes, and step-by-step action labels to represent the scene~\cite{hu2022mile,hu2023uniad}.
Those annotations are costly to acquire and often imperfect, creating fundamental scaling limitations: for instance, the autonomous driving dataset nuScenes~\cite{caesar2020nuscenes} required over 100,000 manual object annotations. Such a bottleneck has motivated research to develop more scalable frameworks.

Vision-action models are scalable when they leverage large quantities of data without requiring proportional human effort. That typically involves learning from raw demonstrations without frame-by-frame labels (e.g., only recording human actions) or utilizing self-supervised objectives that forgo manual annotation. A seminal example is Video PreTraining (VPT)~\cite{baker2022vpt}, which learns directly to act from unlabeled video demonstrations of gameplay, scaling to over 70,000 hours of data that would be prohibitively expensive to annotate manually.

An alternative is weak supervision at scale by leveraging internet data. Vision-Language Models (VLMs) like CLIP~\cite{radford2021clip} and SigLIP~\cite{whai2023siglip} learn robust visual representations from readily available image-text pairs, eliminating the need for costly pixel-wise semantic annotations. Recent works such as LLaVA~\cite{liu2023llava} or PaliGemma~\cite{beyer2024paligemma} follow that framework by building on top of pre-trained LLMs, resulting in vision encoders with strong semantic understanding. Building on that foundation, Vision-Language-Action (VLA) Models like RT-2~\cite{zitkovich2023rt2}, OpenVLA~\cite{kim2024openvla} or $\pi0$~\cite{black2024pi0} demonstrate how web-scale knowledge can be transferred to action generation in a robotic context.

In autonomous driving, the shift towards scalability is particularly evident.
Early end-to-end approaches like TransFuser~\cite{Chitta2023TransFuser} or MILE~\cite{hu2022mile} required extensive annotation to learn visual representations (e.g., HD maps, bounding boxes of agents, Bird's-eye-view or camera semantic masks).
Methods such as LMDrive~\cite{shao2024lmdrive}, GPT-DRIVER~\cite{mao2023gptdriver}, LLM-driver~\cite{chen2024llmdriver} or Language-MPC~\cite{sha2023languagempc} leverage pre-trained LLMs, but they all expect to know the position of all agents in the scene at train or inference time.
That implicitly assumes a perfect upstream perception stack, which is often unrealistic.
Newer methods, instead, leverage VLMs to operate directly on the raw visual from expert demonstrations. For instance, DriveGPT4~\cite{xu2024drivegpt4} uses a pre-trained CLIP~\cite{radford2021clip} encoder and a LLaMA-2~\cite{touvron2023llama2} LLM to interpret driving scenes without requiring dense semantic labels.
CarLLaVA~\cite{renz2024CarLLaVA} further demonstrates how foundation models (LLaVA~\cite{liu2023llava} and LLaMA~\cite{touvron2023llama}) enables closed-loop driving from raw sensor data and sparse navigational inputs.

\subsection{Evaluation}

\paragraph{Video generation:} The rapid progress in large-scale generative models enabled the creation of visually compelling and temporally consistent videos~\cite{sora,veoteam2024veo2,chefer2025videojam,gao2024vista,hu2023gaia1,hassan2024gem}.
Those advancements suggest that generative models can capture meaningful representations of the world, potentially serving as `world models' that simulate and understand physical interactions.
Current practices in evaluating generative video models as world models predominantly rely on metrics such as the Fréchet Inception Distance (FID)~\cite{heusel2017fid} and its temporal extension, the Fréchet Video Distance (FVD)~\cite{Unterthiner2019FVD}.
Those metrics, however, focus on perceptual quality rather than task-specific performance.
In addition, they measure the similarity between generated and actual video distributions, assuming they were Gaussian, resulting in very coarse estimations of generative capabilities.
They fall short, for example, in evaluating the model's understanding of physical laws and real-world dynamics necessary for effective world simulation~\cite{zhu2024soraworldsim}.
To address those limitations, new benchmarks such as Physics-IQ~\cite{motamed2025physicsIQ} or PhyWorld~\cite{kang2024phyworld} test models' comprehension of physical principles, including fluid dynamics and solid mechanics.
Those evaluations reveal that despite achieving visual realism, models often lack the understanding to accurately predict physical interactions, highlighting a significant gap in their ability to function as reliable world simulators.
\paragraph{Open-loop driving:} Open-loop evaluation assesses a system's performance by comparing its predicted future trajectories against pre-recorded expert driving behavior. While that method enables evaluation with realistic traffic data without simulation, it has key limitations: most critically, it fails to measure performance in the actual deployment distribution, which comprises `reasonable' trajectories that deviate from the expert's~\cite{codevilla2018offlineEval, dauner2023parting}. Such limitation is further highlighted in AD-MLP~\cite{zhai2023ADMLP}, which demonstrates that a simple MLP model taking only ego-motion as input may achieve comparable or better open-loop scores than complex perception-based methods.
Furthermore, the distance between predicted and recorded trajectories is not an ideal metric in multi-modal scenarios; for instance, when merging into a turning lane, both immediate and delayed merging could be valid options, but open-loop evaluation penalizes the option not observed in the data~\cite{zhai2023ADMLP}.
To address those limitations, some metrics propose to cover more comprehensive aspects such as traffic violations, progress, and driving comfort~\cite{dauner2023parting}, but the correlation between open-loop performance and actual driving performance remains loose~\cite{dauner2023parting}. 

\paragraph{Closed-loop driving:} Closed-loop driving evaluation addresses the key limitations of open-loop testing by enabling model decisions to influence subsequent observations. Existing approaches can be categorized based on their simulation capabilities and trade-offs. Bird's-eye-view-only simulators~\cite{Gulino2023waymax, caesar2021nuplan} focus on trajectory planning but cannot evaluate end-to-end perception-based systems. Meanwhile, synthetic simulators such as CARLA~\cite{dosovitskiy2017carla} enable comprehensive end-to-end evaluation with dynamic agents, but their synthetic nature introduces significant domain gaps when transferring to real-world scenarios~\cite{chen2023e2esurvey}. Vista~\cite{amini2020vista,amini2022vista2} attempts to bridge this gap through view reprojection from actual data but cannot simulate dynamic agent interactions. NavSim~\cite{Dauner2024navsim} introduces a non-reactive paradigm where agents commit to actions based on initial real-world sensor input and continue the simulation in BEV. That limits long-horizon evaluation as the system does not receive environmental feedback.
Conversely, NeuroNCAP~\cite{ljungbergh2024neuroncap} currently represents the most complete solution by enabling closed-loop evaluation with dynamic agents and continuous sensory feedback through neural rendering, thus allowing for long-horizon scenarios while maintaining photorealism from real-world data.

%% file: tex/method.tex
\section{Models}
\label{sec:method}

\begin{figure}[t]
    \centering
    \includegraphics[width=0.9\linewidth]{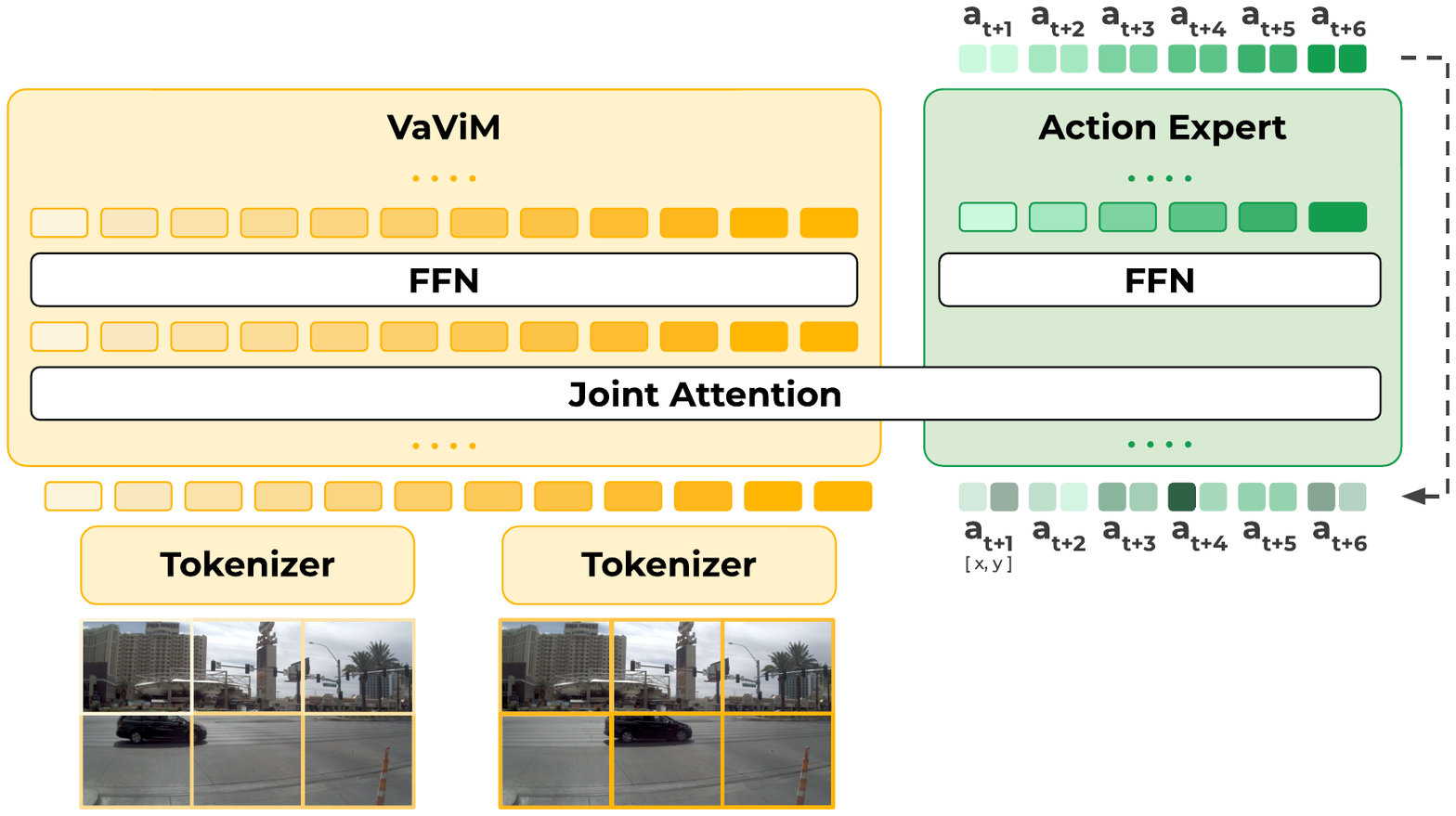}
    \caption{\textbf{End-to-end pipeline of \vam{}.} 
    From a context of up to 8 frames, first \vm (in yellow) builds a spatio-temporal representation, and then \vam{}'s action expert (in green) estimates the dynamic profile of the driving actions to undertake, as a trajectory of 6 waypoints sampled at 2 Hz.}
    \label{fig:vam_overview}
\end{figure}

Our video-action model \vam{} is composed of \vm{} --- a self-supervised video model that learns semantic driving features through next-token prediction (\autoref{sec:model:vm}) --- and an action expert module (\autoref{sec:model:vam}) that enables end-to-end autonomous driving from video inputs.
\autoref{fig:vam_overview} illustrates that perception-to-action pipeline.

\subsection{\vm{} = Auto-regressive Video-Model on tokenized video stream}
\label{sec:model:vm}

At its core, the auto-regressive video model captures the underlying dynamics of driving scenes by modeling the joint distribution of spatio-temporal token sequences. It operates on discrete video tokens, i.e., compact representations of video frames obtained through an image tokenizer (\autoref{sec:model:vm:tokenizer}). By learning to predict the next token in these sequences (\autoref{sec:model:vm:autoregressive}), \vm{} builds a rich understanding of the temporal patterns in driving environments.

\subsubsection{Image Tokenizer}
\label{sec:model:vm:tokenizer}

The architecture starts by transforming continuous image data into a discrete sequence of tokens using vector quantization, a process known as visual tokenization \citep{van2017vqvae}, by mapping local image features to their nearest neighbors in a learned codebook. The codebook acts as a compressed visual vocabulary, enabling efficient auto-regressive modeling while preserving the essence of the visual information for downstream tasks.

More formally, consider a video clip with $T$ frames, with each frame $X_t \in \mathbb{R}^{h \times w \times c}$ for $t \in \{1, \dots, T\}$. Here, $h \times w$ is the spatial resolution, and $c$ is the number of channels. The encoder $f_{\theta} : X \rightarrow e$ processes each frame independently to produce a latent embedding $e \in \mathbb{R}^{h' \times w' \times d}$. For a given frame embedding $e$, at each spatial location $(i,j) \in h' \times w'$, we quantize $e^{(i,j)}$ to a discrete token $q^{(i,j)}$ by performing a nearest-neighbor lookup in the codebook $\{e_k\}$:

\begin{equation}
q^{(i,j)} := \arg \min_k \| e^{(i,j)} - e_k \|_2.
\label{eq:nn-lookup}
\end{equation}

In this equation, $e_k$ represents the embedding vectors in a shared codebook of size $\mathbb{R}^{K \times d}$, where $K$ is the number of discrete entries or codebook vectors, and $d$ is the dimensionality of each vector. The discrete token map $q$ is then used to retrieve the corresponding embeddings, resulting in the embedding map $e_q$.

To map the tokenized representation back into the image domain, we use a decoder $g_{\theta}$, which takes the embedding map $e_q$ and generates the reconstructed output $\hat{x} := g_{\theta}(e_q)$.
That process resembles a standard autoencoder but includes a unique non-linearity that maps latent representations to one of $K$ embedding vectors.

During training, the codebook is initialized randomly and jointly optimized with the encoder and decoder. That adapts the codebook to the data distribution, capturing the most relevant visual features for the task.

The image tokenizer is trained using a vector quantization objective, combining reconstruction, commitment, and adversarial losses to ensure high-fidelity and perceptually realistic image reconstructions. A straight-through estimator computes the gradients, thus handling the non-differentiable nearest-neighbor lookup. For detailed formulations, see \cite{sun2024llamagen}.

\subsubsection{Auto-regressive next token predictor}
\label{sec:model:vm:autoregressive}

The second stage of our approach generates videos in the latent space of the pre-trained Vector Quantized Variational AutoEncoder (VQ-VAE) tokenizer. We use an auto-regressive framework inspired by Large Language Model (LLM) pre-training, employing a transformer decoder to predict tokens sequentially. That allows the model to generate video content patch by patch, capturing spatial and temporal dependencies.

\paragraph{Objective:} 
The model learns the conditional probability of each token given its preceding tokens. For a sequence of $n$ tokens $ \mathcal{Q} = [q^{0}, q^{1}, \dots, q^{n-1}]$, the joint distribution is factorized as the product of conditional probabilities:
\begin{equation}
    P(\mathcal{Q}; \theta) = \prod\limits_{i=1}^{n} P(q^{i}|q^{0}, q^{1}, \dots, q^{i-1}; \theta)
    \label{eq:ar_modeling}
\end{equation}
where $\theta$ are the model parameters. 

We train the model to minimize the negative log-likelihood of the observed token sequences:
\begin{equation}
    \mathcal{L}_{\theta} = -\sum_{i=1}^n \log P(q^{i}|q^{0}, q^{1} \dots, q^{i-1}; \theta)
    \label{eq:NLL}
\end{equation}

We use a softmax function on the model's logits to produce a probability distribution over the vocabulary. We train the model using teacher forcing with cross-entropy loss, aligning the predicted and true token distributions.

\paragraph{\vm{}: model architecture}  
Building upon the previously described tokenizer discretization, we employ a GPT-2~\cite{radford2019language} architecture to model the temporal dynamics of video tokens auto-regressively (following \autoref{eq:ar_modeling}).  While the tokenizer's vocabulary focuses on the perceptual compression of individual frames, our model learns a new set of embeddings optimized for capturing spatio-temporal relationships in the token sequence. Those embeddings map the tokenizer's discrete codes $q$ into a continuous latent space $z$ where spatio-temporal relationships can be modeled auto-regressively.

At each layer $l \in L$, where $L$ is the total number of layers, the computation is as follows:

\begin{align}
z & \leftarrow z + \text{CausalAttn}(\text{LN}(z))  \\ 
z & \leftarrow z + \text{FFN}(\text{LN}(z)).    
\end{align}

where \( \text{FFN}(\cdot) \) denotes a fully connected layer, \( \text{CausalAttn}(\cdot) \) is a causal attention layer with masking~\cite{waswani2017attention}, and \( \text{LN}(\cdot) \) denotes layer normalization~\cite{lei2016layer}.
We use GELU~\cite{hendrycks2016gaussian} as the activation function and employ weight tying between the input embedding layer and the output projection layer to reduce the number of parameters. Additionally, at inference time, a KV cache~\cite{ott2019fairseq} is maintained for efficient auto-regressive sampling.

Following GAIA-1~\cite{hu2023gaia1}, we use two types of learned positional embeddings to capture both spatial and temporal dependencies. The \emph{spatial positional embedding}  is shared across all frames, allowing the model to capture spatial dependencies within each image independently. In contrast, the \emph{temporal positional embedding} is unique for each frame, enabling the model to capture dependencies across frames in a video. By combining these two positional embeddings, the model effectively learns both intra-frame and inter-frame relationships.

\subsection{\vam{} = \vm{} + action expert}
\label{sec:model:vam}

\begin{figure}[h]
\centering
\begin{subfigure}{0.45\linewidth} 
    \includegraphics[clip, width=\linewidth]{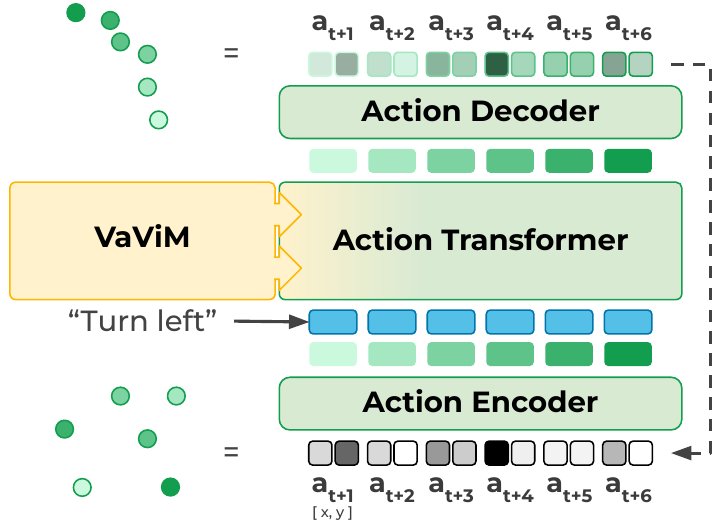}
      \caption{\textbf{Flow-matching action expert}}
    \label{fig:actionexpert}
\end{subfigure}
\hspace{1.5cm}
\centering
\begin{subfigure}{0.35\linewidth}
    \includegraphics[clip, width=\linewidth]{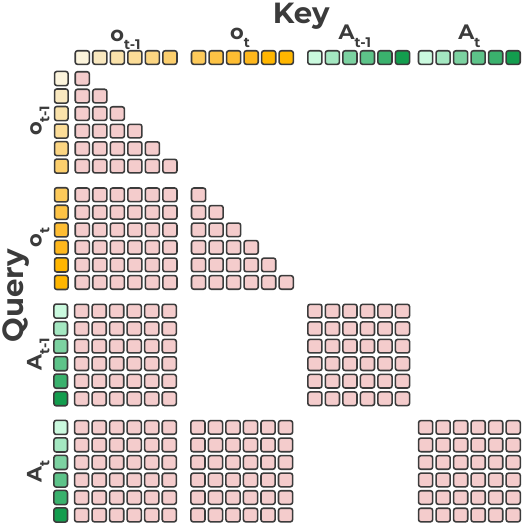}     
      \caption{\textbf{Joint Attention Masking}}
    \label{fig:joint_attention_mask}
\end{subfigure}
\caption{\textbf{Model details.}  (\subref{fig:actionexpert}) The iterative denoising process for driving trajectory estimation: starting from random noise, \vam estimates the sequence of driving waypoints (green dots), conditional to high-level commands (e.g., `turn left') and \vm{} features.  (\subref{fig:joint_attention_mask}) The joint attention between \vm{} tokens $o$ and action tokens $A$ at training time.}
\label{fig:actionexpert_system}
\end{figure}

Whether video generation pre-training effectively captures the features essential for safe and reliable driving is a key question. To bridge the gap between pre-trained video representations and driving decisions, we introduce an action expert module, forming \vam{} by complementing \vm{} with decision-making.
The action expert, inspired by $\pi0$~\cite{black2024pi0}, uses flow matching to generate actions by progressively denoising a noisy ego-trajectory, illustrated on the bottom left of~\autoref{fig:actionexpert}, into a coherent driving trajectory. The denoising is conditioned on high-level driving commands (e.g., `turn left', `go straight') and video features from \vm{} encoding the scene dynamics.
While $\pi0$ conditions on single frames for robotic manipulation, we extend it to driving by exploiting the temporal contexts of multiple frames that are crucial for understanding dynamic scenarios.

We adopt flow matching instead of alternatives such as action quantization~\cite{lee2024vqbet} because it directly learns the vector fields that define the transport to the target probability distribution, enabling accurate action generation while effectively modeling complex multimodal distributions. That is particularly important for our trajectory dataset, which is challenging to capture with quantization-based methods due to its long-tail distribution dominated by straight trajectories, with maneuvers such as U-turns appearing rarely (\autoref{fig:trajectories}). 

More formally, we assume a dataset of driving recordings and their associated high-level commands $\mathcal{D} = \{(O_t, A_t, c_t)\}$, with $O_t = [o_{t}, \dots, o_{t-N}]$ representing a sequence of images observed up to the $N$ past frames; $c_t \in \{\text{left}, \text{right}, \text{straight}\}$ being the high-level commands, which act as a guide for the vehicle direction, \eg, `turn left', on~\autoref{fig:actionexpert}; and $A_t = [a_{t+1}, \dots, a_{t+H}]$ being the `action', defined as a sequence of $[x, y]$ ego-positions in the BEV reference-frame specifying the dynamic profile of the driving path to undertake. We illustrate the `action' trajectory at the top left of~\autoref{fig:actionexpert}. The `action' is extracted from the pre-recorded ego-motion over the next $H$ future timesteps after the current timestep $t$.

Through the combination of \vm{} and the action expert module, \vam effectivelly learns the conditional vector fields $v_\theta(A^{\tau}_t; O_t, c_t)$ that transport actions sampled from a noised distribution $A^{\tau}_t=[a^{\tau}_{t+1}, a^{\tau}_{t+2}, \dots, a^{\tau}_{t+H}]$ to actions $A_t$ from the observed distribution $O_t$ and high-level commands $c_t$. That denoising process is formalized in \autoref{sec:imitationlearning}.

\paragraph{Architecture.}
In more detail, the action expert consists of an action encoder, a joint attention transformer, and an action decoder (\autoref{fig:actionexpert}).

\begin{itemize}
    \item \textbf{Action Encoder:} It projects the actions into a latent space using an MLP and incorporates positional embeddings of the flow matching step $\tau$, a learned temporal embedding (``action at time $t$'') and a learned embedding for each high-level command (left, right, straight).
    
    \item \textbf{Joint Attention Transformer:} This module enables interaction between action representations and visual features, conditioning the denoising process on observed scene dynamics coming from \vm{}. We use a specialized attention masking scheme illustrated in~\autoref{fig:joint_attention_mask}
    \begin{itemize}
        \item Action tokens attend to all past context frames and all other action tokens within the same frame. 
        \item Visual tokens maintain causal masking to preserve their sequential nature, preventing them from being conditioned by future observations. 
    \end{itemize}
    
    \item \textbf{Action Decoder:} It maps the latent action features back to the action space with a linear layer, predicting the denoising vector field $v_\theta(A^{\tau}_t; O_t, c_t)$.
\end{itemize}

The architecture provides two key advantages. First, \vm{} and the action expert interact exclusively through joint attention. This design choice allows the action expert to use a smaller MLP dimensionality than \vm{} while maintaining matching dimensions in attention layers. Such dimensional reduction is crucial for efficiency, as the action expert performs multiple forward passes during iterative denoising and sampling. Second, the layer-wise joint attention addresses the challenge of feature extraction from \vm{}'s layers. Different layers capture varying levels of abstraction—from raw vocabulary embeddings to task-specific features. Rather than selecting and committing to a single layer, the joint attention mechanism learns to extract relevant features across \vm{}'s entire depth.

During inference, we sample trajectories by integrating the denoising vector field over 10 steps using the forward Euler method, starting from random noise $A^0_t \sim \mathcal{N}$. That integration process progressively refines the noisy actions into a coherent driving trajectory that satisfies both the high-level command and environmental constraints captured by the temporal à

%% file: tex/data.tex
\section{Data and Training}

\subsection{Data}
\label{sec:data}

Our desiderata for the data were to find a large dataset of non-annotated data for the pre-training and a sufficient amount of annotated data (with trajectories synchronized with perception) for fine-tuning. To that end, we train \vm and \vam on a collection of three datasets: OpenDV~\cite{yang2024opendv}, a massive non-annotated web dataset, and nuPlan~\cite{caesar2021nuplan} and nuScenes~\cite{caesar2020nuscenes}, dedicated automotive datasets captured with multiple sensors.

\input{figures/datasets}

\noindent\paragraph{OpenDV~\cite{yang2024opendv}} The OpenDV dataset, illustrated on~\autoref{fig:opendv_examples}, is the largest driving dataset publicly available up to now, with more than 1700 hours of driving videos, collected at 60 FPS, resulting in over 360 million frames. The dataset comprises single-camera front-cam videos collected from YouTube, with annotated durations of intros (usually 90 seconds) and outros (usually 60 seconds) for trimming, to avoid artifacts such as title sequences and closing credits. Most videos are at or close to Full HD (1920$\times$1080) resolution.

We include in our data only the videos at exactly Full HD to avoid issues of aspect ratio distortion. That meant discarding 1.3\% of the videos (2.5\% of the total duration). Using FFMPEG~\cite{tomar2006converting}, we extracted the frames for the remaining videos at 10 FPS and 512$\times$288 pixels, discarding intros and outros. We stored the frames in individual JPEG files. 
We extract overlapping clips of 8 frames at 2 FPS to train \vm. Only a front camera is available for this dataset, without any metadata.

\noindent\paragraph{nuPlan~\cite{caesar2021nuplan}} The nuPlan dataset contains around 1200 hours of driving scenarios recorded in Las Vegas (838 hours), Boston, Pittsburgh, and Singapore. In particular, among the 1200-hour raw data, approximately 94-hour recordings contain sensor information (LiDAR and cameras) with a sampling rate of 10 Hz. Our project only employs the RGB images in 1274 recorded videos and the ego position. More specifically, we only use the front camera instead of the eight cameras that cover the 360-degree view around the ego vehicle illustrated in \autoref{fig:nuplan_examples}. We collect from nuPlan 2,833,723 frames for training and 492,477 for validation, together with trajectory extracted from the ego positions. We also extract overlapping clips of 8 frames at 2 Hz from the original 10 Hz video sequences.

\noindent\paragraph{nuScenes~\cite{caesar2020nuscenes}} The nuScenes dataset contains 1000 driving scenes of 20 seconds collected in Boston and Singapore in~\autoref{fig:nuscenes_examples}. Similarly to nuPlan, while nuScenes includes 6 cameras, LIDAR, RADAR, \etc, we restrict its usage to the front camera and ego position in this work. That results in a dataset of 28,130 training frames and 6,019 validation frames. We also extract overlapping clips of 8 frames. The dataset is natively synchronized at 2Hz.

As detailed in subsequent sections, we use the datasets for different steps of \vm and \vam training. Specifically, we use OpenDV only for \vm pre-training~\autoref{subsec:muP}. A mix of the three datasets to fine-tune \vm~\autoref{subsec:finetuning}. For both training steps, only the front camera is used, making those steps completely unsupervised and, thus, highly scalable. Finally, we use nuScenes and nuPlan to learn \vam through imitation learning on the ego trajectory~\autoref{sec:imitationlearning}, which are illustrated on~\autoref{fig:trajectories}. That training stage requires access to the ground-truth expert trajectory, although, interestingly, recent approaches such as GEM~\cite{hassan2024gem}, explore the use of pseudo-annotations for the ego trajectory, paving the way to scaling action learning to OpenDV-size datasets.

\input{figures/trajectories_datasets}

\subsection{\vm{} pre-training}
\label{subsec:muP}

Training large autoregressive models requires careful consideration of both parameterization and scaling strategy. In this section, we present our approach to efficiently scale \vm{} beyond 1 billion parameters.
\paragraph{Compute-Efficient Scaling with $\mu$P:} 
We adopt the Maximal Update Parametrization ($\mu$P)~\cite{yang2022miup} to enable efficient scaling, inspired by recent advances in large-scale model training~\cite{hu2024minicpm,dey2023cerebras,yao2023nanolm,li2023flm101b}. $\mu$P allows us to use the same learning rate and optimization hyperparameters across different model scales, simplifying hyperparameter tuning and stabilizing training dynamics.

$\mu$P works by reparameterizing the network's initialized weights, activations, and learning rate on a per-layer basis proportional to the network width. That enables width-independent updates and zero-shot hyperparameter transfer as model width scales~\cite{lingle2024largescalemiup, yao2023nanolm}\footnote{\href{https://blog.eleuther.ai/mutransfer/}{Eleuther.ai --- muP Practitioner's Guide}}\footnote{\href{https://cloneofsimo.notion.site/What-to-do-to-scale-up-09e469d7c3444d6a90305397c38a46f5}{cloneofsimo --- What to do to scale up?}}.

We initiate hyperparameter optimization with a base model width of 256 (60M parameters). Using approximately 20\% of the pre-training data, we conduct a grid search over 50 randomly sampled hyperparameter configurations, requiring around 900 GPU hours. This phase establishes the core training hyperparameters. We then scale up to a model with width 768 (185M parameters) to fine-tune the weight decay parameter, with each candidate configuration requiring about 1254 GPU hours. At this point, all hyperparameters are fixed.

\begin{figure}[h]
    \centering
    \includegraphics[width=0.6\linewidth]{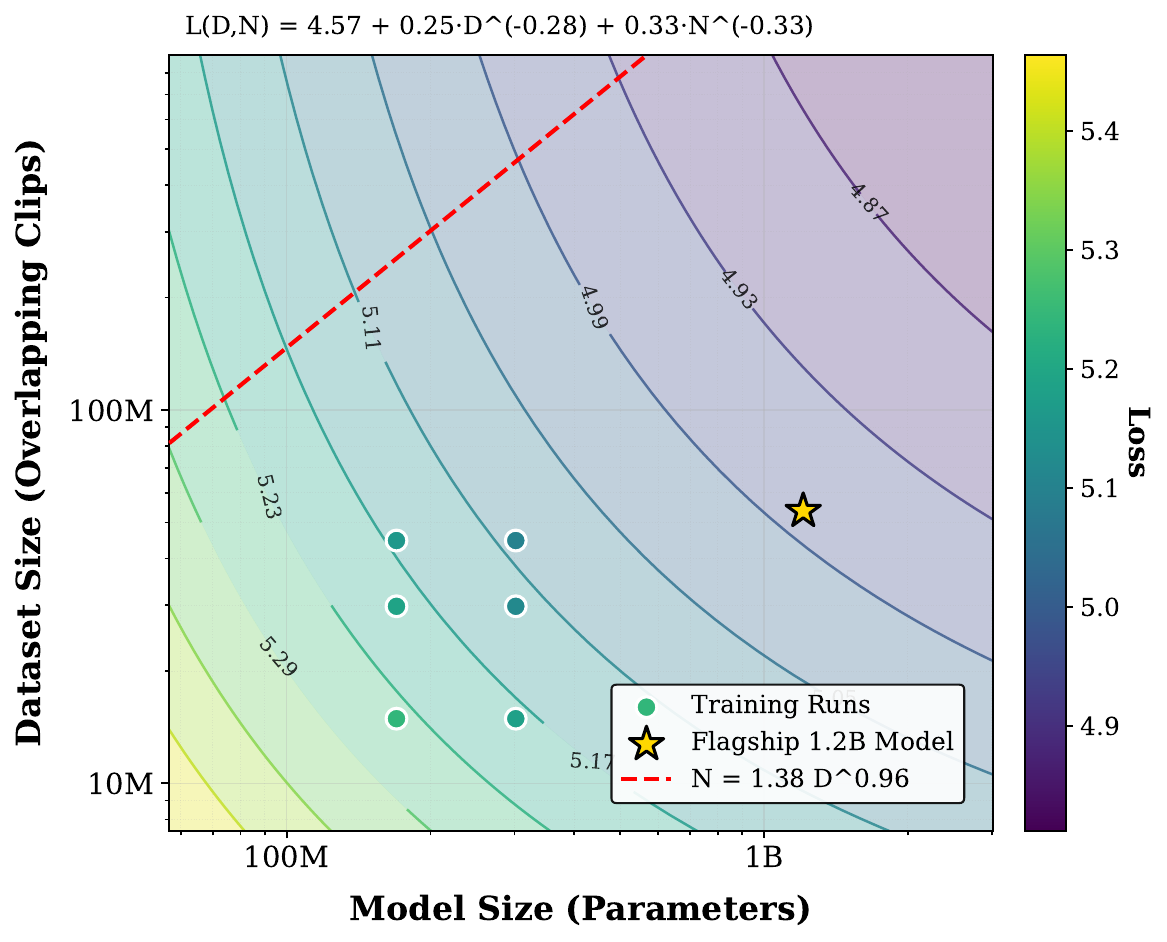}
    \caption{\textbf{\vm{} scaling law} predicts the expected validation loss (next-token cross-entropy) of the 1.2B \vm{}-L model and indicates that more data would strongly benefit our models. The green dots correspond to checkpoints used to fit the scaling law; the yellow star is the predicted performance of \vm{} (0.06\% error); the dotted red is the compute-optimal frontier~\cite{hoffmann2022chinchilla}.}
    \label{fig:scaling_law}
\end{figure}

\paragraph{Establishing Scaling Laws:} 
A scaling law empirically models the validation loss as a power law of the training data and model size. Following~\cite{hoffmann2022chinchilla}, we define it as:
\begin{equation}
     L(D, N) = L_0 + A \cdot D^{-\alpha} + B \cdot N^{-\beta},
\end{equation}

where:
\begin{itemize}
    \item $L(D, N)$ is the validation loss as a function of the dataset size $D$ and model size $N$.
    \item $D$ is the number of video clips for training the model.
    \item $N$ is the number of non-embedding model parameters.
    \item $L_0$ represents the irreducible error of an ideal model.
    \item $A$, $\alpha$, $B$, and $\beta$ are positive constants that capture the sensitivity to data and model size.
\end{itemize}

We estimate the parameters of the scaling law ($L_0$, $A$, $B$, $\alpha$, and $\beta$) by training a model with width 1024 (318M parameters), consuming approximately 1762 GPU hours. We use the Warmup-Stable learning rate schedule~\cite{hu2024minicpm} to extract multiple pre-trained checkpoints within one training run. Specifically, we save checkpoints at 25\%, 50\%, and 75\% of the pre-training data for both width 768 and 1024 models.

Combining those checkpoints across different model and data scales, we fit an empirical scaling law that captures performance improvements with increasing data and model size. Using SciPy's~\cite{virtanen2020scipy} \emph{minimize} with the L-BFGS-B~\cite{nocedal1980LBFGS,zhu1997LBFGSB} optimizer, we find the optimal parameters: $(L_0, A, \alpha, B, \beta) = (4.574, 0.250, 0.279, 0.326, 0.329)$. That allows us to derive the optimal scaling relationship: $D = 1.384 \cdot N^{0.962}$.

We show the scaling law isoplot and its derived optimal scaling schedule in~\autoref{fig:scaling_law}. Our analysis reveals that despite using the large-scale OpenDV dataset \citep{yang2024opendv} with more than 1700 hours of driving data, our model is under-trained, suggesting that additional data could significantly improve performance. 
In total, we use 10,186 GPU hours to establish the scaling laws before the learning-rate decay phase, including 8,424 GPU hours to finalize the search of all hyperparameters. We then train the flagship model with 1.2B parameters and observe an estimation error of only 0.003 (~0.06\%) compared to the predicted scaling law.

\subsection{Fine-tuning \vm{} on target dataset}
\label{subsec:finetuning}

Building upon our pre-trained model, we implement a carefully designed fine-tuning strategy that leverages multiple datasets to enhance the model's performance. Our approach begins with a checkpoint selected from the `stable' phase of pre-training.

For the fine-tuning phase, we construct a diverse training mix combining three complementary datasets: (1) OpenDV, a large-scale, diverse dataset that provides broad coverage of general driving scenarios across the world; (2) nuPlan, a more specialized dataset that aligns with our subsequent imitation learning phase; (3) nuScenes, that serves the dual purpose of supporting imitation learning and targetting the NeuroNCAP evaluation, \ie, the target task of driving. 

From OpenDV, we initially extracted around 59M overlapping video clips and allocated 90\% of them for pre-training (warmup and stable learning rate phases) and reserved the remaining 10\% for fine-tuning (learning rate decay phase). However, rather than utilizing the entire fine-tuning portion, we strategically sample from multiple sources to create a balanced training mix:

\begin{itemize}[leftmargin=1cm]
    \setlength{\itemsep}{0.5em}
    \item 40\% from OpenDV -- 2,385,300 clips
    \item 58.72\% from nuPlan -- 2,765,278 clips, representing the complete nuPlan dataset
    \item 1.28\% from nuScenes -- 76,120 clips, achieved by repeating the nuScenes dataset four times
\end{itemize}

We leave for future works to evaluate the impact of only using the 10\% portion of OpenDV. An additional open question is the optimal composition of the fine-tuning data mix, possibly with different proportions of the datasets above or additional autonomous driving datasets.

\subsection{Training \vam{} with imitation learning}
\label{sec:imitationlearning}

A carefully structured imitation learning allows transforming our pre-trained video model (\vm{}) into an actionable video-action model (\vam{}). This section outlines how we enable end-to-end driving capabilities while preserving the rich visual representations learned during pre-training.

As discussed in~\autoref{sec:model:vam}, our approach employs flow matching, building upon the framework introduced in $\pi$0~\cite{black2024pi0}. More formally, given a dataset of expert demonstrations with associated high-level commands $\mathcal{D} = \{(O_t, A_t, c_t)\}$, with $O_t = [o_{t}, \dots, o_{t-N}]$ representing the sequence of images observed up to N past frames, the high-level command $c_t \in \{\text{left}, \text{right}, \text{straight}\}$ and the expert trajectory $A_t = [a_{t+1}, \dots, a_{t+H}]$ of future positions over horizon $H$, we learn to denoise trajectories through a conditional probability flow.

The key insight of flow matching lies in its elegant formulation of the forward process and induced vector field. We learn a conditional denoising vector field $v_\theta$, which defines how to progressively transform noisy trajectories back into expert-like behavior. The training process follows a forward noising schedule defined by:

\begin{equation}
    A^\tau_t = \tau A_t + (1-\tau)\epsilon, \quad \epsilon \sim \mathcal{N}(0, I)
    \label{eq:noised_action}
\end{equation}

That process represents a linear interpolation between the expert action $A_t$ and Gaussian noise $\epsilon$. The variable $\tau \in [0,1]$ represents the noise level. This process smoothly interpolates between expert actions ($\tau = 0$) and pure noise ($\tau = 1$) and traces out paths in the action manifold. For training, the action expert uses the following objective to predict the denoising vector field $v_\theta$:

\begin{equation}
    L^\tau(\theta) = \mathbb{E}_{p(A_t|O_t, c_t),q(A^\tau_t | A_t)} ||v_\theta(A^\tau_t, O_t, c_t) - u(A^\tau_t|A_t)||^2
\end{equation}

where $q(A^\tau_t|A_t)$ is the forward process defined above and $u(A^\tau_t|A_t)$ is the optimal transport vector field. The optimal transport vector field $u(A^\tau_t|A_t)$ represents the ideal direction in which noisy actions should move to become expert actions. Our learned vector field $v_\theta$ approximates this optimal transport.
The vector field acts as the generator of a continuous transformation on the manifold of plausible driving actions. It generates a flow that transforms a simple distribution (Gaussian noise) into our target distribution of expert actions.
During inference, we generate action sequences by integrating the learned vector field:
\begin{equation}
A^{\tau+\delta}_t = A^\tau_t + \delta \cdot v_\theta(A^\tau_t, O_t, c_t)
\end{equation}
using 10 steps of the forward Euler method, starting from random noise $A^0_t \sim \mathcal{N}(0, I)$.

That framework enables our model to capture complex multimodal action distributions directly from the expert demonstration. The effectiveness of this approach is extensively demonstrated in \autoref{sec:evaluation_actionexpert}, where we show strong performance in both open-loop prediction and closed-loop driving scenarios.

\subsection{Implementation and Training details}

\paragraph{Tokenizer.}
We use a pre-trained image tokenizer, LlamaGen~\cite{sun2024llamagen}, which is based on the VQGAN architecture. Specifically, we use the stride=16 tokenizer, which has 72M parameters. It has a vocabulary size of 16,384 with codewords of 8 dimensions. We use images of size 512$
\times$288, resulting in a token map of 32$
\times$18, or 576 tokens.

\paragraph{\vm} is based off a GPT-2 transformer architecture~\cite{radford2019language}. We train it with a context length of 8 frames, resulting in a maximum context of 4,608 tokens. It has 24 layers, a vocabulary size of 16,384, with a width scaling from 768 (\vm-S) to 1024 (\vm-B), up to 2048 (\vm-L). This results in a codebook of size 
 12.6M, 16.8M, and 33.65M respectively. We keep the dimensionality of the heads fixed at 128, making the number of attention heads scale with the model size. We set a standard multiplication factor of 4 for the FFN hidden dimensionality. We optimize it with AdamW~\cite{AdamW}, a base learning rate of 0.0041, a weight decay of 1$e$-7, and $\beta=(0.9, 0.95)$ while clipping the gradient with a norm of 1.0. Finally, we initialize with a standard deviation of 0.0289. As described in~\autoref{subsec:muP}, the $\mu P$ parameterization scales the learning rate per layer according to the width layer (see our code or \cite{yang2022miup} from exact specifications). We train all our models with a batch size of 384 and vary the number of GPUs depending on the model size to maximize GPU utilization.

\paragraph{\vam} predicts the trajectory for the 6 next timesteps at 2 Hz, \ie, for 3 seconds. The dimensionality of \vam attention layers is identical to its \vm companion for the joint attention (\autoref{sec:model:vam}). However, \vam's MLP layers dimensionality is reduced by a factor of 4 with respect to \vm's dimensionality for efficient action sampling; \ie 192 for \vam-S, 256 for \vam-B, and 512 for \vm-L. For the joint attention, we must project the tokens to match the \vm's dimensionality. We use a learning rate equal to 0.0194, an initialization standard deviation of 0.0086, and similar optimizer parameters to \vm{}. For the flow matching loss, we follow $\pi0$ and use a beta distribution for the noise schedule and 10 steps for denoising at inference time. We efficiently train our model with different observation context lengths using a block attention pattern (\autoref{fig:joint_attention_mask}). That allows training the action expert to handle varying lengths of temporal context from one training clip.

\paragraph{Training infrastructure.} We detail the compute requirements for our most compute-intensive run, training \vm-L. We run our training on 48 nodes of 4 H100s, totaling 192 GPUs. To help scale that job, we employ lightning~\cite{Falcon_PyTorch_Lightning_2019} and deepspeed-stage2~\cite{deepspeed}. The total running time for the training is around 25 hours, totaling around 4800 (H100) GPU hours. We pre-train the model on approximately 60 million overlapping windows.

%% file: figures/datasets.tex



\begin{figure}[h]
\centering
\begin{subfigure}{0.3\linewidth} 
\includegraphics[trim={0cm 0cm 15cm 0.0cm},clip, width=\linewidth]{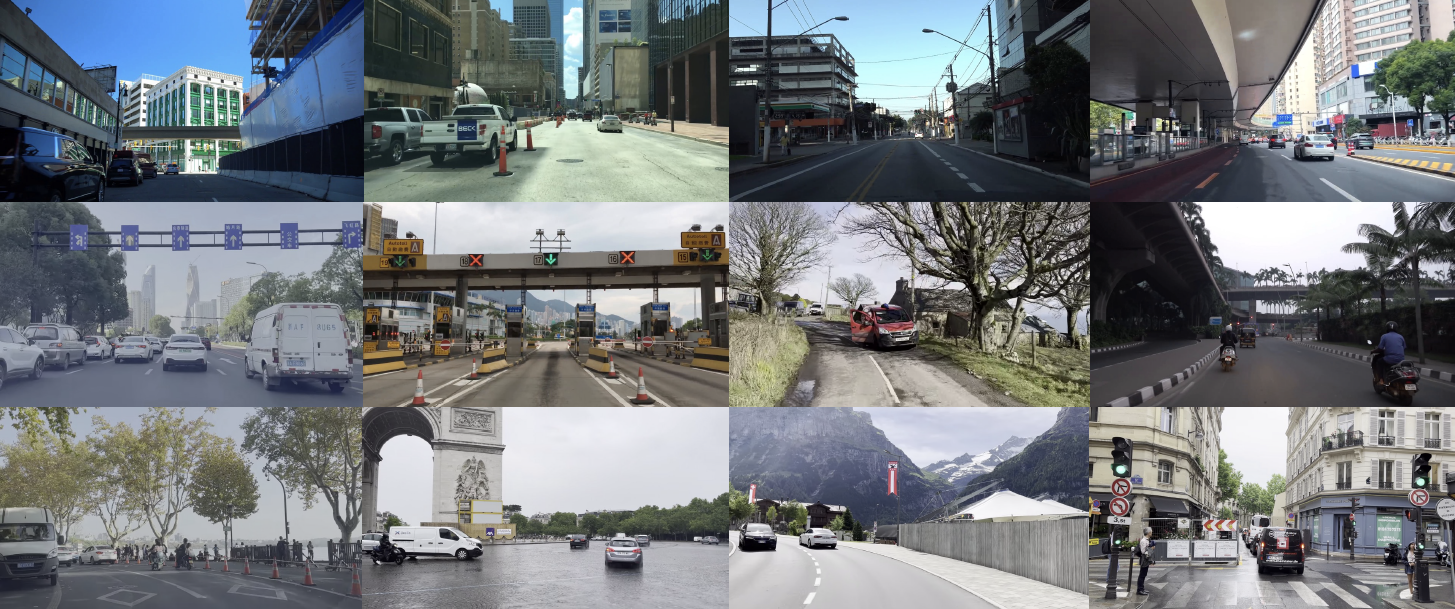}
  \caption{OpenDV~\cite{yang2024opendv}}
\hfill
\label{fig:opendv_examples}
\end{subfigure}
\centering
\begin{subfigure}{0.32\linewidth} 
\includegraphics[trim={0cm 0cm 0 0.0cm},clip, width=\linewidth]{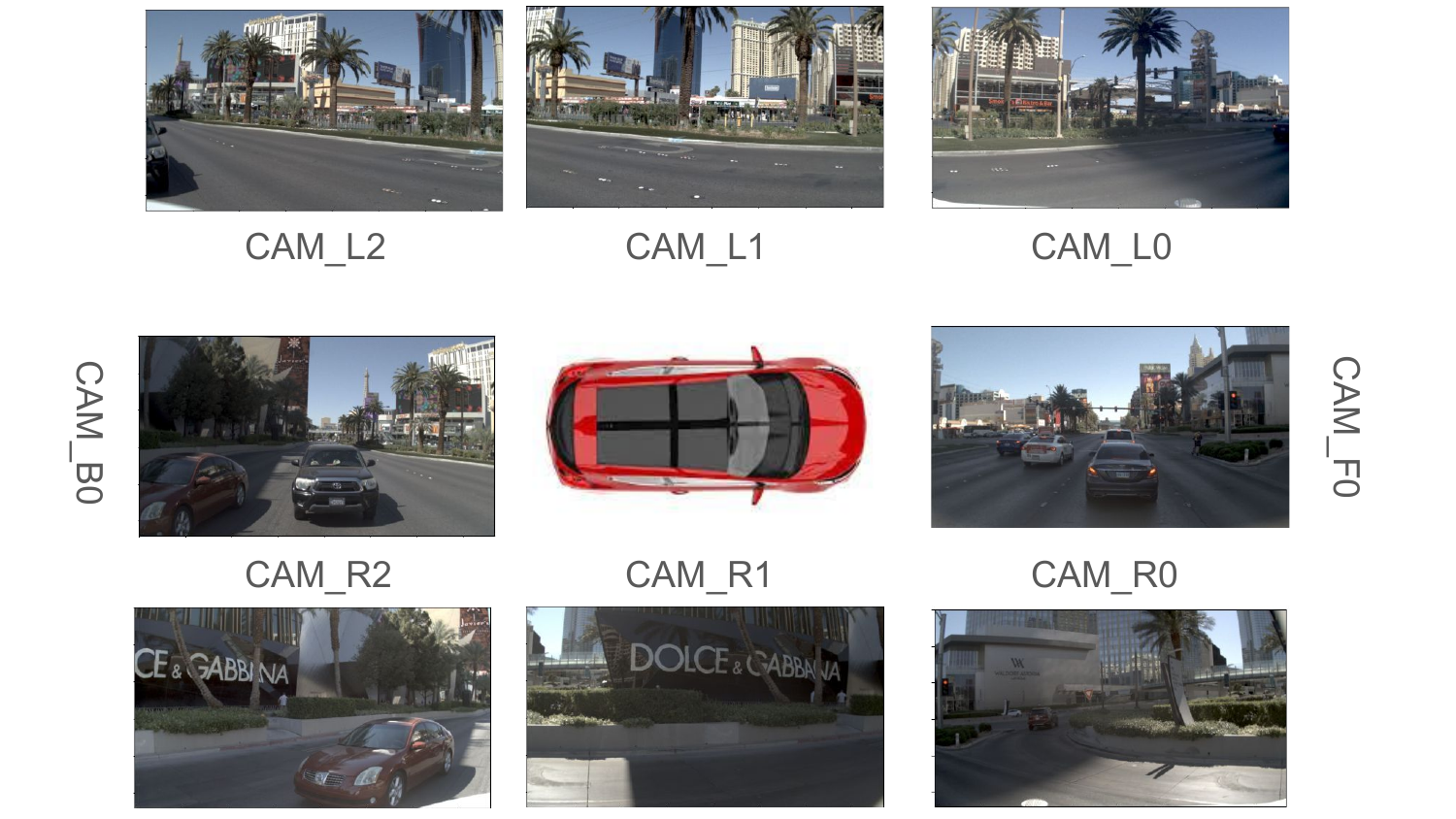}
  \caption{nuPlan~\cite{caesar2021nuplan}}
\hfill
\label{fig:nuplan_examples}
\end{subfigure}
\centering
\begin{subfigure}{0.32\linewidth}
\includegraphics[trim={0cm 0cm 0 0cm},clip, width=\linewidth]{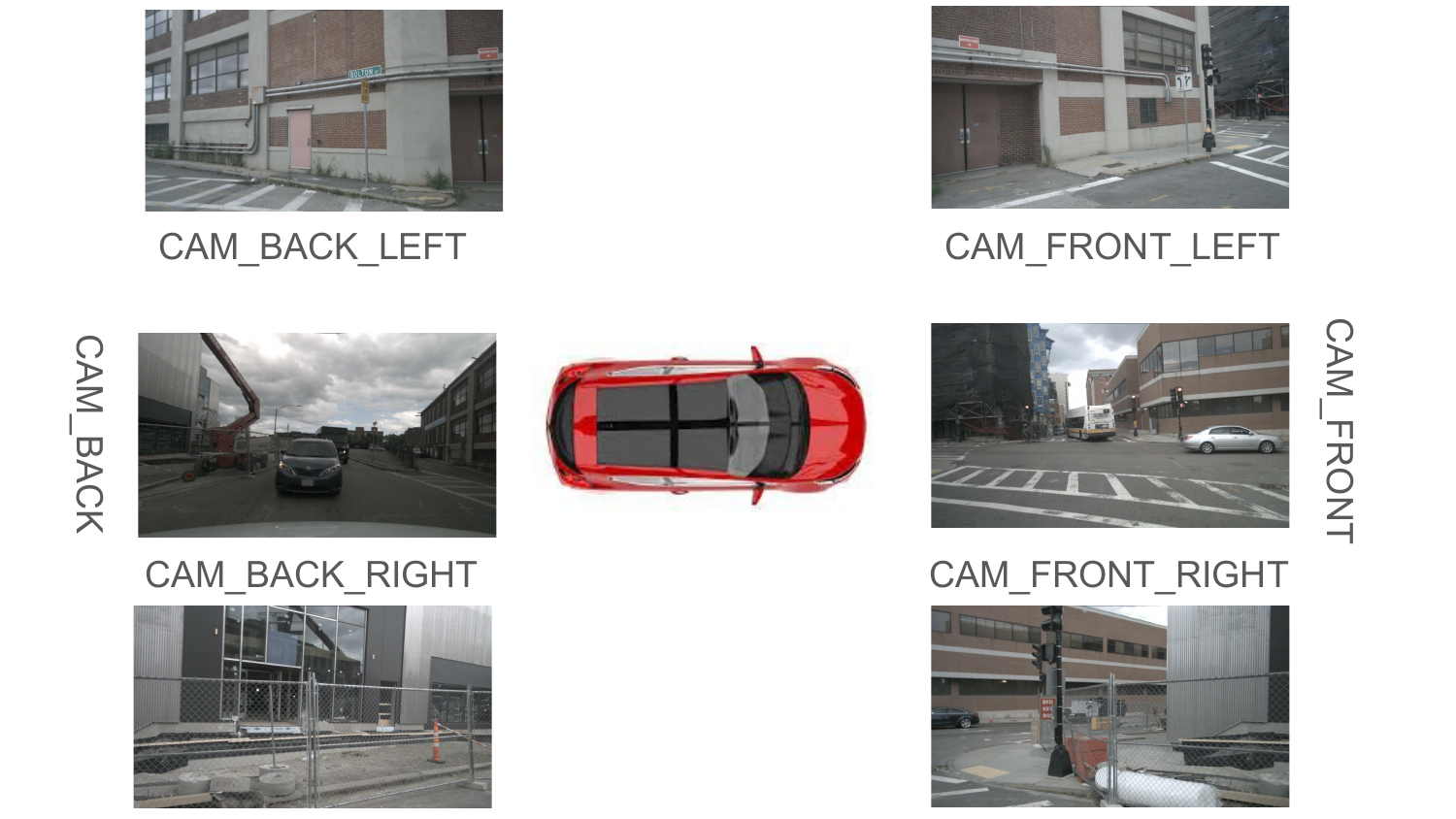}     
  \caption{nuScenes~\cite{caesar2020nuscenes}}
  \hfill
\label{fig:nuscenes_examples}
\end{subfigure}
\caption{\textbf{Image examples.} RGB dash-cam images from (\subref{fig:opendv_examples}) the OpenDV dataset~\cite{yang2024opendv}, (\subref{fig:nuplan_examples}) the eight cameras in nuPlan~\cite{caesar2021nuplan}, and (\subref{fig:nuscenes_examples}) the six cameras in nuScenes~\cite{caesar2020nuscenes}.}
\label{fig:data_examples}
\end{figure}

%% file: figures/trajectories_datasets.tex
\begin{figure}[h]
\centering
\begin{subfigure}{0.48\linewidth} 
\includegraphics[trim={0.3cm 0.7cm 3cm 0.8cm},clip, width=\linewidth]{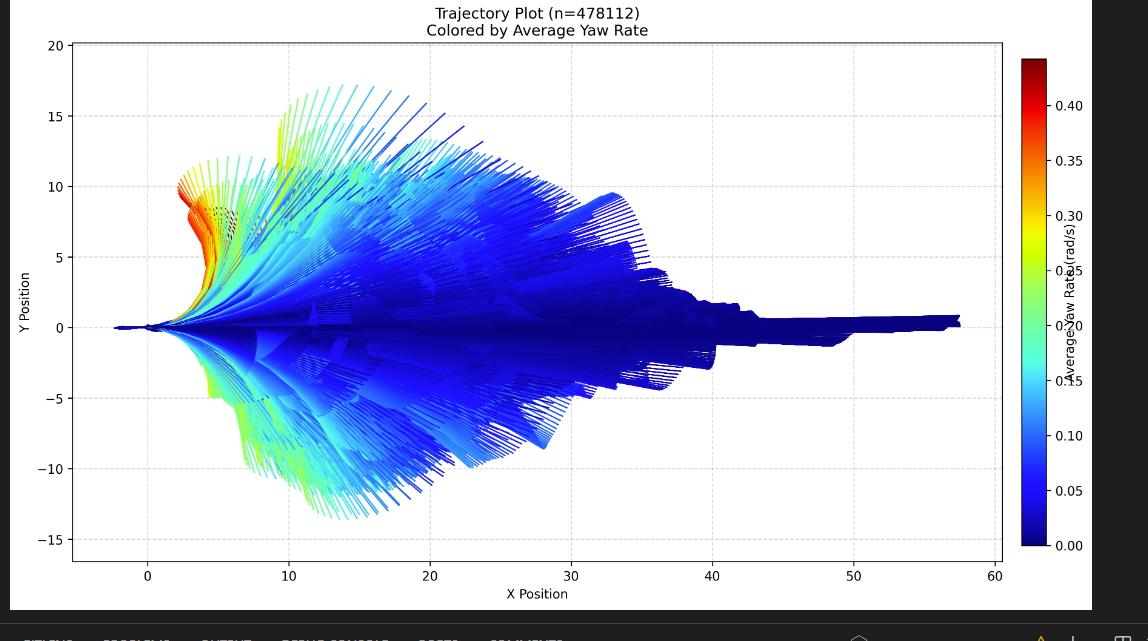}
  \caption{nuPlan~\cite{caesar2021nuplan}}
\hfill
\label{fig:nuplan_trajs}
\end{subfigure}
\centering
\begin{subfigure}{0.48\linewidth}
\includegraphics[trim={0cm 0cm 3cm 1.85cm},clip, width=\linewidth]{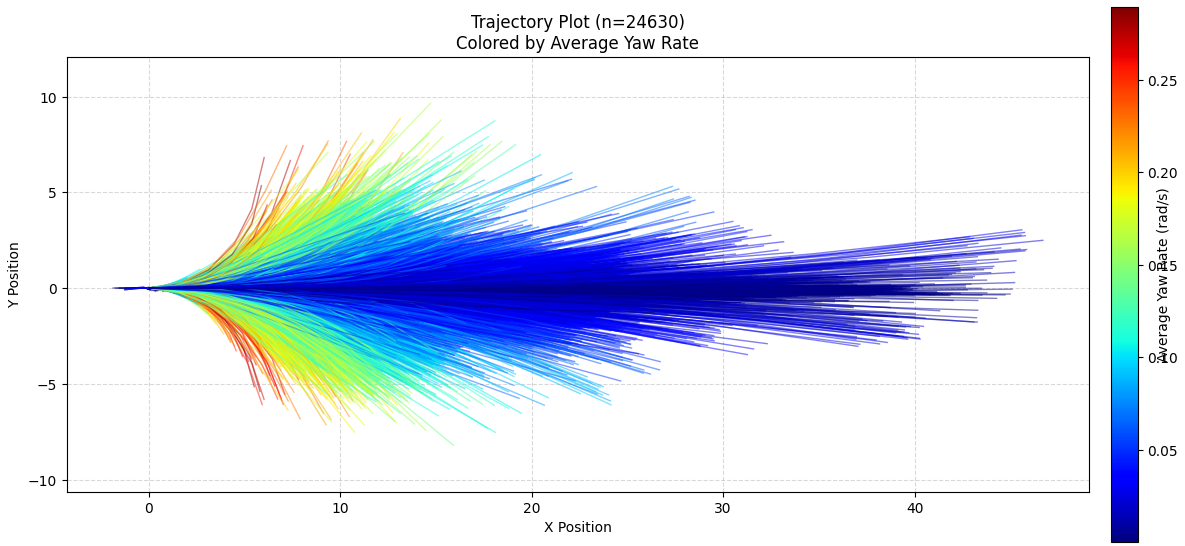}     
  \caption{nuScenes~\cite{caesar2020nuscenes}}
  \hfill
\label{fig:nuscenes_trajs}
\end{subfigure}
\caption{\textbf{Trajectory distributions.} The heading of the ego-car is oriented towards the positive y-axis. The trajectories are colored, per set of trajectories, by average yaw rate (rad/s) from blue as minimum to red as maximum. The maximum distance travelled within 3s is around 60~m in nuPlan and 40~m for nuScenes. The maximum average yaw rate is around 0.40 rad/s in nuPlan~(\subref{fig:nuplan_trajs}) and 0.25 rad/s for nuScenes~(\subref{fig:nuscenes_trajs}).}
\label{fig:trajectories}
\end{figure}

%% file: tex/eval.tex
\section{Experiments and Results}
\label{sec:exps_and_results}

\paragraph{Additional evaluation datasets.} We evaluate \vm{} and \vam on datasets not used for their training, namely Cityscapes~\cite{Cordts2016Cityscapes} and KITTI~\cite{geiger2013kitti}. We refer to the latter as \emph{KITTI} when we downsample the clips to 8 frames at 2 Hz (to match the training FPS) and as \emph{KITTI-1f} when we downsample the clips to a single frame.

\subsection{\vm{} video pre-training evaluation}

\subsubsection{Generation quality}
\input{tables/fid}

To evaluate the quality of the generation of our \vm, we use the Frechet Inception Distance (FID)~\cite{heusel2017fid}. To account for non-object-centric settings typical of driving datasets, we use the features from a DINOv2 model, which have been shown to be richer. 

Specifically, given a context of 4 frames, we generate 4 frames with \vm. We use the features of the 4 context frames as reference images to compute the FID. We compute an FID score for each future frame individually (FID@t), \ie, FID@2 means the FID of the second generated frame. Since \vm does not directly generate images but tokens, we compute the FID for the LLamaGen-VQGAN tokenizer to act as an upper bound, as it serves as an ``oracle'', \ie, it encodes a real frame of the ``future''. We choose the same reference frames for the FID. And compute the FID individually for the 4 ground-truth future frames.
That setting changes slightly compared to the reconstruction FID, as the features of the original images considered for reconstruction are not part of the reference images used to compute the FID.
Note that, although we pre-train on a video clip, we do not use Frechet Video Distance~\cite{Unterthiner2019FVD} (FVD) for evaluation because it relies on an I3D~\cite{I3D} that requires at least 10 frames as input.

From~\autoref{tab:fid}, we observe a clear scaling benefit, where larger models consistently achieve better performance, where \vm-L outperforms \vm-B, which further outperforms \vm-S. That trend highlights the advantage of increasing model capacity to improve reconstruction quality and other evaluation metrics.

We also show qualitative results for generation on~\autoref{fig:video_generation}. We feed a context of 4 frames~\autoref{fig:context_frames} to either \vm-S~\autoref{fig:video_generation_vms} or \vm-L~\autoref{fig:video_generation_vml}. We observe that both models generate frames that are spatially coherent, with correctly generated large structures. Furthermore, we observe that \vm-L predicts frames closer to the true future (car on the left turning right) by inferring the motion from the 4 first frames. That contrasts with the generative behavior of \vm-S.

\input{figures/video_generation}

\subsubsection{Semantic segmentation}

\input{tables/segmentation}

In~\autoref{tab:segmentation}, we evaluate VaViM for semantic segmentation using the Humming-bird approach~\cite{balazevic2023hummingbird}. Specifically, we use the features of layer 12 of our VaViM's transformer to encode a frame. We sample 10 patch-features per image, using the sampling approach of the open-source Humming-bird implementation~\cite{pariza2024hbird}. We report the mean Intersection over Union (mIoU), which measures the overlap between predicted and ground truth segmentations across all classes, providing a comprehensive assessment of per-class and overall segmentation quality. Higher mIoU scores indicate better alignment with ground-truth annotations.

As observed in recent studies~\cite{Empirical_autoregressive}, auto-regressive models struggle compared to discriminative models like Dino and Dinov2~\cite{Caron2021EmergingPI,oquab2024dinov2}. We can see Dino models outperform \vm on all setups. However, this study shows that, even with the auto-regressive pre-training that does not explicitly enforce semantic understanding, \vm still manages to segment the different datasets in a zero-shot manner: indeed, neither Cityscapes nor KITTI was part of the training data mix.
\autoref{tab:segmentation} also shows that during fine-tuning on our target datasets, \vm does not lose the representation capability learned on the diverse pre-training data.

Moreover, we conduct a qualitative study on the features obtained with \vm on~\autoref{fig:pca_qual_results}. As proposed by~\cite{Caron2021EmergingPI}, we visualize as RGB primaries the 3 main components of a PCA of the features. The semantic consistency of \vm's is visible as similar colors being assigned to objects of the same class (\eg, pedestrians, cars, or road), which suggests the features hold semantic meaning, even if they are not invariant enough to perform at semantic segmentation.

\input{figures/pca_qual_results}

\subsection{\vam{} driving evaluation}
\label{sec:evaluation_actionexpert}

\subsubsection{Open-loop evaluation}

In~\autoref{tab:minADE}, we evaluate \vam in an open-loop setup. We compute the $\textbf{minADE}_k$ ($\downarrow$), \ie, the minimum over $k$ sampled trajectories of the Average Distance Error, taken as the average of point-wise L$^2$ distances between the closest (among $k=5$) sampled trajectory and the ground-truth expert trajectory. The metric is calculated for both nuPlan~\cite{caesar2021nuplan} and nuScenes~\cite{caesar2020nuscenes}.


We observe a clear trend where scaling improves minADE, which shows that among the different samples, \vam is able to match the expert ones better when increasing the compute ($\sim \text{\# params} \times \text{\# data}$). Qualitatively, we observe that as smoother trajectories for ``better'' models. However, as evidenced by prior works~\cite{codevilla2018offlineEval,dauner2023parting} and our results in closed-loop evaluation (\autoref{sec:close_loop}), open-loop scores are not predictive of good driving capabilities.

\input{tables/minADE}

\subsubsection{Closed-loop evaluation}
\label{sec:close_loop}

\textbf{NeuroNCAP:} While the previous open-loop evaluation demonstrates strong trajectory prediction accuracy, it fails to capture the cascading effects of the model's decisions. Closed-loop evaluation addresses that limitation by allowing decisions to influence future observations, thus providing a more realistic assessment of safety-critical behavior.

To evaluate our model's performance in closed-loop, we employ NeuroNCAP~\cite{ljungbergh2024neuroncap}, a simulator specifically designed for testing autonomous driving systems in safety-critical scenarios. To the best of our knowledge, it is currently the only existing data-based closed-loop simulator. Other solutions are either synthetic~\cite{dosovitskiy2017carla} (leading to domain gap) or based on view reprojection~\cite{amini2020vista,amini2022vista2} (leading to limited novel views). 

NeuroNCAP employs a NeRf-based simulator that executes the driving model decision and generates the corresponding novel view. That enables photorealistic closed-loop evaluation of driving models. In particular, a key feature of NeuroNCAP is its ability to insert pre-defined adversarial agents into the scene, such as a vehicle following hazardous trajectories. 

Using that capability, the framework creates challenging test conditions inspired by the European New Car Assessment Programme (Euro NCAP), featuring three primary scenario types: stationary obstacles in the ego-lane, frontal collisions with oncoming vehicles, and side collision scenarios from cross-traffic. In our experiments, we leverage this framework to systematically evaluate our \vam model's ability to handle those challenging scenarios while maintaining its intended trajectory. \newline

\noindent \textbf{Collision metrics:} NeuroNCAP's evaluation protocol relies on two metrics: (1) the collision rate as a percentage of scenarios without collision and (2) the NeuroNCAP score (NNS) that assigns scores based on collision avoidance success and impact velocity reduction, offering a quantitative measure of the model's safety performance. More formally:
\begin{equation}
    \text{NNS} = \begin{cases}
        5.0 & \text{if no collision} \\
        4.0 \cdot \max(0, 1 - v_i/v_r) & \text{otherwise}
    \end{cases}
\end{equation}
\newline

\noindent \textbf{Baselines:} We first compare \vam with existing NeuroNCAP baselines. The Base-U and Base-V baselines are naïve methods that use the perception outputs from UniAD~\cite{hu2023uniad} and VAD~\cite{jiang2023vad}, respectively. They operate on a simple rule-based approach: maintaining constant velocity unless an object is detected in a corridor ahead of the ego-vehicle ($\pm$2 meters laterally and up to 2$*v_{ego}$ meters longitudinally). If an object is detected within this corridor (TTC $<$ 2s), they initiate a braking maneuver. The second class of baselines are state-of-the-art end-to-end planners, UniAD~\cite{hu2023uniad} and VAD~\cite{jiang2023vad}, which process 360° camera input, CAN-bus signals, and high-level commands to predict future trajectories.

\input{tables/neuro_ncap}

A quantitative comparison appears in \autoref{tab:eval-WM}. Traditional baseline models establish a strong foundation in static scenarios, with Base-U and Base-V achieving NeuroNCAP scores of 4.72 and 4.82, respectively, setting a robust performance threshold for advanced approaches to surpass.

VAD demonstrates superior overall performance, achieving a 12\% reduction in collision rates compared to our approach across all test conditions. Our qualitative assessment reveals that while \vam effectively identifies and responds to dynamic obstacles, it seldom initiates complete stops or urgent braking, even in scenarios where such actions would be optimal, such as encountering a stationary bus in the middle of the road.

Neither \vm nor \vam predicts occupancy, which was shown to boost performances when used to post-process estimated trajectories~\cite{ljungbergh2024neuroncap}, by allowing optimization with a classical solver. Such post-processing might also benefit \vam. Moreover, \vam currently relies exclusively on front-cam input, giving it an inherent disadvantage against systems with 360 camera arrays, especially for side-impact scenarios. Despite those limitations, \vam demonstrates competitive capabilities, notably surpassing UniAD's in side scenarios by 4\%. In frontal scenarios, \vam achieves state-of-the-art performance, with a NeuroNCAP score of 2.38.

\noindent \textbf{Limitations of the benchmark:} We unintentionally produced a noisy model that avoided the road, as it never encountered the scripted hazardous scenarios, it achieved exceptionally high safety scores in the benchmark ($>$ 4.0 NNS). That finding highlights a critical gap in the evaluation framework: while it effectively measures collision avoidance, it does not adequately assess adherence to intended driving behavior or route completion metrics.

\input{tables/neuro_ncap_frontal_vam}

To address those limitations and provide a more comprehensive evaluation, we propose to use two complementary metrics. First, we measure the mean deviation from the guiding trajectory, which quantifies how well the model adheres to intended driving paths. It differs from the ADE metrics by measuring the mean instantaneous distance to the closest point of the reference trajectory instead of ADE's pairwise distance. Second, we introduce a goal-progress metric that measures the relative reduction in distance to the destination, formally defined as:

\begin{equation}
    \text{progress\_toward\_goal} = \max(0.0, \frac{d_{\text{initial}} - d_{\text{final}}}{d_{\text{initial}}})
    \label{eq:progress}
\end{equation}

where $d_{\text{initial}}$ and $d_{\text{final}}$ represent the initial and final distances to the goal, respectively. That formulation ensures the metric remains bounded between 0 and 1, where 1 indicates complete goal achievement. Together, those metrics effectively capture both the quality of trajectory following and task completion, providing a more robust framework for evaluating autonomous driving systems that prevent the gaming of safety metrics through undesirable driving behaviors.

\begin{wrapfigure}{L}{0.5\textwidth}
    \vspace{-1em}
    \centering
    \includegraphics[width=\linewidth]{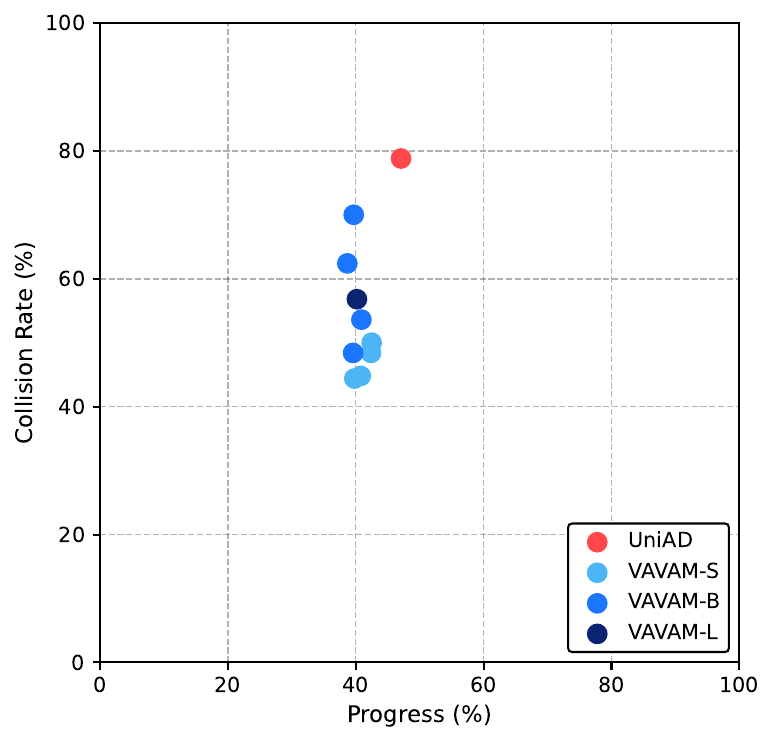}
    \caption{\textbf{Collision Rate vs. Progress.} An ideal model would be positioned at the bottom right corner of this plot, attaining the defined goal position without any collisions. \vam{} makes progress towards that ideal by being safer than UniAD, with 27\% less collisions, with minimal degradation to progress.}
    \label{fig:collision_rate_vs_progress}
    \vspace{-1em}
\end{wrapfigure} 

Using those metrics, we study the behavior of \vam in \autoref{tab:ncap_frontal}. We focus on the score of the frontal scenario as our model is front-cam only. We plot the \emph{collision rate} vs \emph{progress} in \autoref{fig:collision_rate_vs_progress}. Although our model significantly reduces collision rate, it only achieves comparable \emph{collision rate} to UniAD. We expected that as we scale the model and training data, our model would be able to increase the \emph{progress} score while improving metrics related to collisions (NeuroNCAP score and collision rate). However, as we scale, whether in data or model size, the \emph{progress} metric varies unpredictably, and the \emph{collision rate} tends to increase. At the same time, we observe that the \emph{mean deviation} decreases.

We hypothesize that such a phenomenon is mainly due to the limitations of the imitation learning methodology and the definition of the high-level command. Indeed, the model is trained exclusively through imitation learning on pre-recorded expert trajectories, which serve a dual purpose: as training data and as a guiding signal during training and inference.

During training, the model learns to follow exactly the trajectories presumably executed safely by expert drivers. However, at test time, when facing an adversarial vehicle, the model must simultaneously respect that learned behavior (following the trajectory) while adapting to a situation that may require significant deviation from it (avoiding collision).

As the training compute scales up, collision rate increases and the mean deviation metric decreases, which suggests that larger and more trained models may be overfitting to the trajectory-following behavior. Rather than learning the underlying decision-making process that still allows for safe deviation, the model becomes more rigid in its adherence to the guiding trajectory. 


%% file: tables/fid.tex
\begin{table*}
    \caption{
    Evaluation of video generation quality on KITTI~\cite{geiger2013kitti} and nuScenes~\cite{caesar2020nuscenes}. FID@t is the generation FID on the \textit{t}-th frame (lower is better). Larger models tend to have better FIDs. The FIDs are within the ballpark of the upper bound given by the `oracle' implemented with LlamaGen~\cite{sun2024llamagen}.
    } 
    \centering
    \setlength{\tabcolsep}{3pt}
    \resizebox{\columnwidth}{!}{
    \begin{tabular}{l c cccc cccc}
        \toprule
         \multirow{2}{*}{\Th{Model}} & \multirow{2}{*}{\Th{\# params}}  &  \multicolumn{4}{c}{\Th{KITTI} \citep{geiger2013kitti}} &  \multicolumn{4}{c}{\Th{nuScenes} \citep{caesar2020nuscenes}} \\
         \cmidrule(lr){3-6} \cmidrule(lr){7-10}
          && \Th{FID}@1 & \Th{FID}@2 & \Th{FID}@3 & \Th{FID}@4 & \Th{FID}@1 & \Th{FID}@2 & \Th{FID}@3 & \Th{FID@4}  \\
         \midrule
         \multicolumn{10}{c}{\cellcolor{valeocell} {\texttt{Oracle upper-bound}}} \vspace{0.3em} \\
         \Th{LlamaGen} & - & 6.8 & 7.4 & 7.7 & 8.1 & 9.8 & 9.9 & 10.2 & 10.4 \\
         \midrule
         \multicolumn{10}{c}{\cellcolor{valeocell} {\texttt{Fine-tuned}}} \vspace{0.3em} \\
         \Th{\vm{}-S} & 185M & 15.1 & 22.1 & 28.3 & 34.5 & 23.9 & 30.8 & 35.8 & 39.8  \\  
         \Th{\vm{}-B} & 318M & 13.7 & 19.5 & 24.6 & 30.0 & 20.7 & 27.5 & 33.2 & 38.6  \\ 
         \Th{\vm{}-L} & 1.2B & 12.0 & 16.4 & 20.8 & 24.7 & 16.8 & 19.6 & 23.2 & 26.1 \\
         \bottomrule
    \end{tabular}
    }
    \label{tab:fid}
\end{table*}

%% file: figures/video_generation.tex
\begin{figure}[h]
\centering
\begin{subfigure}{\linewidth} 
\includegraphics[width=\linewidth]{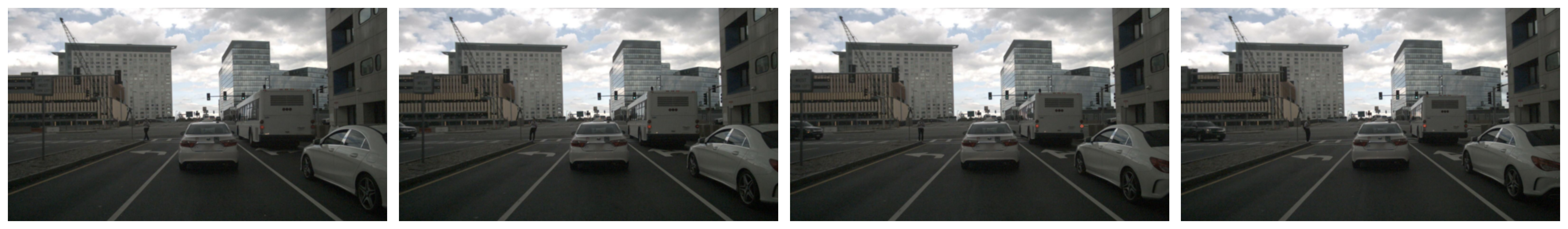}
\caption{Real context frames.}
\label{fig:context_frames}
\end{subfigure}
\begin{subfigure}{\linewidth} 
\includegraphics[width=\linewidth]{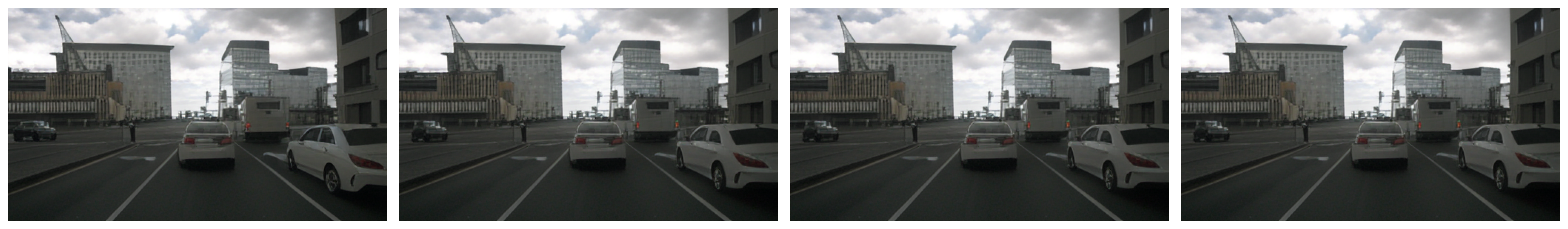}
\caption{Generated frames with \vm-S (fine-tuned).}
\label{fig:video_generation_vms}
\end{subfigure}
\begin{subfigure}{\linewidth} 
\includegraphics[width=\linewidth]{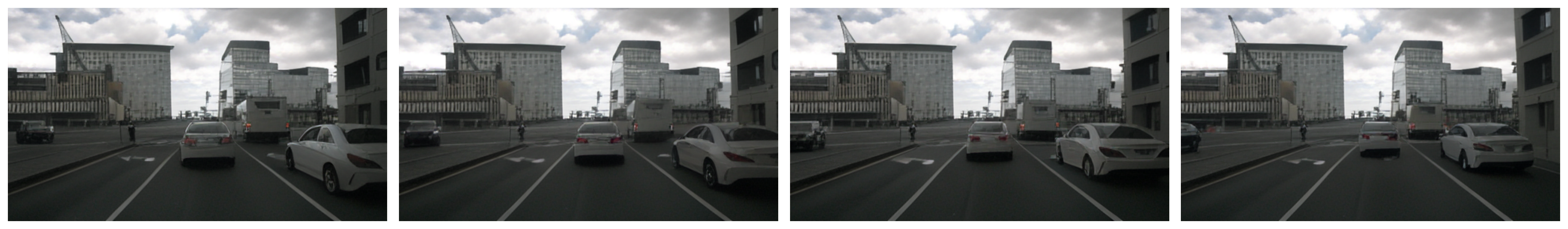}
\caption{Generated frames with \vm-L (fine-tuned).}
\label{fig:video_generation_vml}
\end{subfigure}
\caption{\textbf{Video generation with \vm}. Given 4 context frames (\subref{fig:context_frames}), we generate the next 4 frames with \vm-S (\subref{fig:video_generation_vms}) and \vm-L (\subref{fig:video_generation_vml}). Both models correctly predict the overall structure, but only \vm-L generates the expected motion.}
\label{fig:video_generation}
\end{figure}

%% file: tables/segmentation.tex
\begin{table*}
    \caption{Evaluating \vm{} on zero-shot semantic segmentation with Humming-Bird~\cite{balazevic2023hummingbird} on Cityscapes~\cite{Cordts2016Cityscapes} and KITTI~\cite{geiger2013kitti}. Results in mIoU (higher is better).}
    \centering
    \begin{tabular}{l l c c c}
        \toprule
         \multirow{1}{*}{\Th{Model}} & \multirow{1}{*}{\# \Th{params} (in M)} &  \multicolumn{1}{c}{\Th{Cityscapes} \citep{Cordts2016Cityscapes}} &  \multicolumn{1}{c}{\Th{KITTI-1f} \citep{geiger2013kitti}} &  \multicolumn{1}{c}{\Th{KITTI} \citep{geiger2013kitti}} \\
         \midrule
         \multicolumn{5}{c}{\cellcolor{valeocell} {\texttt{{Baseline Models}}}} \vspace{0.3em}\\
         \Th{Dinov1-B}~\cite{Caron2021EmergingPI} & 85   & 31.5   & 34.7   & 35.1 \\
         \Th{Dinov2-B}~\cite{oquab2024dinov2} & 86   & 43.4   & 41.8   & 41.6  \\
         \Th{Dinov2-L}~\cite{oquab2024dinov2} & 300  & 13.0   & 16.5   & 16.7 \\
         \Th{Dinov2-g}~\cite{oquab2024dinov2} & 1100 & 37.8   & 40.9   & 40.6 \\
         \midrule
          \multicolumn{5}{c}{\cellcolor{valeocell} {\texttt{{Pre-trained}}}} \vspace{0.3em}\\
         \Th{\vm{}-S} & 185      & 20.0   & 25.5   & 22.1 \\ 
         \Th{\vm{}-B} & 318      & 21.2   & 25.1   & 23.3 \\ 
         \Th{\vm{}-L} & 1200     & 17.9   & 21.0   & 20.4 \\
         \midrule
          \multicolumn{5}{c}{\cellcolor{valeocell} {\texttt{{Fine-tuned}}}} \vspace{0.3em}\\
         \Th{\vm{}-S} & 185      & 20.3   & 25.4   & 23.2 \\
         \Th{\vm{}-B} & 318      & 20.8   & 25.3   & 23.4 \\
         \Th{\vm{}-L} & 1200     & 18.4   & 22.0   & 20.3 \\
         \bottomrule
    \end{tabular}
    \label{tab:segmentation}
\end{table*}

%% file: figures/pca_qual_results.tex



\begin{figure}[h]
\centering
\begin{subfigure}{0.48\linewidth} 
\includegraphics[width=\linewidth]{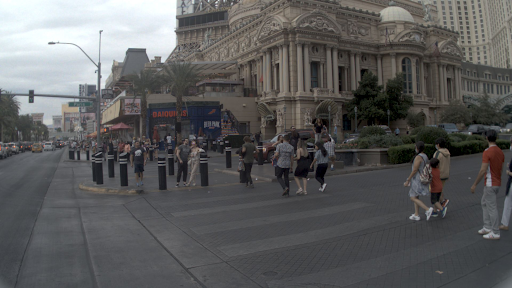}
\hfill
\end{subfigure}
\centering
\begin{subfigure}{0.48\linewidth} 
\includegraphics[width=\linewidth]{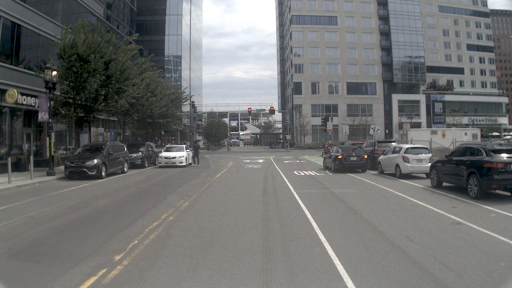}
\hfill
\end{subfigure}
\centering
\begin{subfigure}{0.48\linewidth} 
\includegraphics[width=\linewidth]{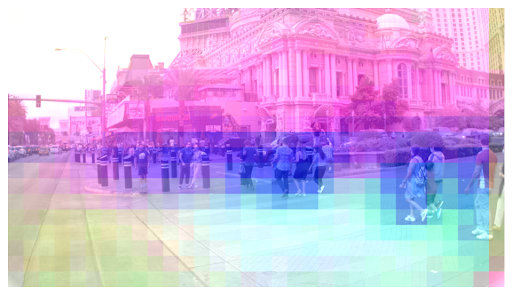}
\hfill
\end{subfigure}
\centering
\begin{subfigure}{0.48\linewidth} 
\includegraphics[width=\linewidth]{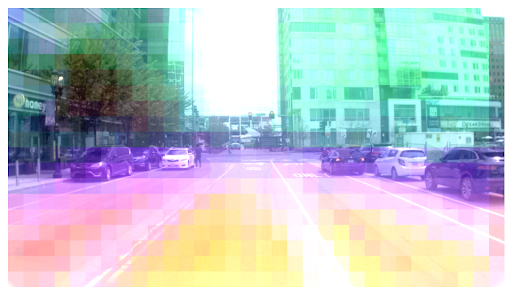}
\hfill
\end{subfigure}
\centering
\caption{\textbf{Main components of \vm's representation.} We take features from \vm{}'s 22nd layer, project them with PCA, and map the three main components to the RGB primaries. The features have a considerable semantic grouping, shown as objects of the same type (pedestrians, cross-walk, road markings, cars, etc.) being assigned similar colors.}
\label{fig:pca_qual_results}
\end{figure}

%% file: tables/minADE.tex
\begin{table*}[t]
    \caption{Open-loop evaluation of \vam measured with minADE$_k$ $(\downarrow)$, with $k=5$. Increasing compute (bigger models, more data) results in a lower minADE. The parameter counts of our models appear as <parameters in \vm{}> + <parameters in the action expert of \vam>.}
    \centering
    \begin{tabular}{l l l c c}
        \toprule
         \multirow{1}{*}{\Th{Model}} & \multirow{1}{*}{\# \Th{params (in M)}} & \multirow{1}{*}{\# \Th{data } (in $\times 10^3$)}  &  \multicolumn{1}{c}{\Th{nuScenes} \citep{caesar2020nuscenes}} &  \multicolumn{1}{c}{\Th{nuPlan} \citep{caesar2021nuplan}} \\
         \midrule
         \multicolumn{5}{c}{\cellcolor{valeocell} \texttt{Video-action models trained from raw data}} \vspace{0.3em} \\
         \multirow{4}{*}{\Th{\vam{}-S}}   & \multirow{4}{*}{185 + 21} & 38   & 1.14  & 0.76  \\ 
         && 77   & 1.13  & 0.83   \\ 
         && 116  & 1.04  & 0.77   \\ 
         && 139  & 1.00  & 0.68   \\ 
         \midrule
         \multirow{4}{*}{\Th{\vam{}-B}} & \multirow{4}{*}{318 + 38} & 38   & 0.96   & 0.57   \\ 
         && 77   & 0.93   & 0.59   \\ 
         && 116   & 0.95   & 0.64   \\ 
         && 139   & 0.85   & 0.53   \\ 
         \midrule
         \Th{\vam{}-L}   & 1,200 + 150   & 139 & 0.80   & 0.52   \\
         \bottomrule
    \end{tabular}
    \label{tab:minADE}
\end{table*}

%% file: tables/neuro_ncap.tex
\newcommand{\cmark}{\ding{51}}
\newcommand{\xmark}{\ding{55}}

\begin{table*}[t]
    \setlength{\fboxsep}{1pt} 
    \caption{Performance on Neuro-NCAP benchmark. \vam{} obtains SOTA scores on the frontal scenarios, despite using only a front-cam and no human annotations. We highlight the post-processing step and the side scenario \colorbox{th}{in gray} to emphasize that \vam{} does not use post-processing of the trajectories~\cite{hu2023uniad} and only uses the frontal camera as input, while other methods have 360° perception. Scores with $^\dagger$ were reproduced by us.
    }
    \centering
    \scriptsize
    \begin{tabularx}{\linewidth}{l c YYYY YYYY}
        \toprule
         \multirow{2}{*}{\Th{model}} & \multirow{2}{*}{\Th{Post-proc.}} & \multicolumn{4}{c}{\Th{NeuroNCAP Score} $\uparrow$} & \multicolumn{4}{c}{\Th{Collision rate} (\%) $\downarrow$} \\
         \cmidrule(lr){3-6} \cmidrule(lr){7-10}
          && \Th{Avg.} & \Th{Stat.} & \Th{Frontal} & \Th{Side} & \Th{Avg.} & \Th{Stat.} & \Th{Frontal} & \Th{Side} \\
         \midrule
         \multicolumn{10}{c}{\cellcolor{valeocell} \texttt{Baseline Model trained w/ hand-labeled annotations}} \vspace{0.3em} \\
         \Th{Base-U} & - & 2.65 & 4.72 & 1.80 & 1.43 & 69.90 & 9.60 & 100.00 & 100.00 \\
         \Th{Base-V} & - & 2.67 & 4.82 & 1.85 & 1.32 & 68.70 & 6.00 & 100.00 & 100.00 \\
         \midrule
         \Th{UniAD} & \xmark & 0.73 & 0.84 & 0.10 & 1.26 & 88.60 & 87.80 & 98.40 & 79.60 \\
         \Th{VAD} & \xmark & 0.66 & 0.47 & 0.04 & 1.45 & 92.50 & 96.20 & 99.60 & 81.60 \\
         \midrule
         \Th{UniAD} & \cmark & 1.84 & 3.54 & 0.66 & 1.33 & 68.70 & 34.80 & 92.40 & 78.80 \\
         \Th{UniAD$^\dagger$} & \cmark & 2.08 & 3.58 & 1.18 & 1.48 & 61.1 & 31.2 & 78.8 & 73.2 \\
         \Th{VAD} & \cmark & \textbf{2.75} & \textbf{3.77} & 1.44 & \textbf{3.05} & \textbf{50.70} & \textbf{28.70} & 73.60 & \textbf{49.80} \\
         \midrule
         \multicolumn{10}{c}{\cellcolor{valeocell} \texttt{Video-action model trained from raw data}} \vspace{0.3em} \\
         \Th{\vam{}-L} & \cellcolor{th}\xmark & 2.46 & 3.55 & \textbf{2.38} & \cellcolor{th}1.47 & 57.90 & 41.40 & \textbf{56.80} & \cellcolor{th}75.60 \\
         \bottomrule
    \end{tabularx}
    \label{tab:eval-WM}
\end{table*}

%% file: tables/neuro_ncap_frontal_vam.tex

\begin{table*}[t]
    \caption{Extended NeuroNCAP scores for the frontal scenario using mean deviation and progress (\autoref{eq:progress}) as additional metrics. We reproduce the scores of UniAD~\cite{hu2023uniad}. The parameter counts of our models appear as <parameters in \vm{}> + <parameters in the action expert of \vam>.}
    \centering
    \begin{tabularx}{\linewidth}{l l l YYYY}
        \toprule
        \multirow{1}{*}{\Th{Model}} & \multirow{1}{*}{\# \Th{params (in M)}} & \multirow{1}{*}{\# \Th{data} (in K)} & \Th{NNS} $\uparrow$ & \Th{Coll. rate} (\%) $\downarrow$ & \Th{Mean Deviation} $\downarrow$& \Th{Progress} (\%) $\uparrow$ \\
        \midrule
        \multicolumn{7}{c}{\cellcolor{valeocell} \texttt{Baseline Model trained w/ hand-labeled annotations}} \vspace{0.3em} \\
        UniAD \citep{hu2023uniad} & - & - & 1.18 & 78.8 & 1.077 & 47.1 \\
        \midrule
        \multicolumn{7}{c}{\cellcolor{valeocell} \texttt{Video-action models trained from raw data}} \vspace{0.3em} \\
        \multirow{4}{*}{\Th{\vam{}-S}} & \multirow{4}{*}{185 + 21} & 38 & 3.019 & 44.4 & 2.616 & 39.8 \\
        && 77 & 3.022 & 44.8 & 2.486 & 40.8 \\
        && 116 & 2.795 & 48.4 & 2.394 & 42.4 \\
        && 139 & 2.669 & 50.0 & 2.022 & 42.5 \\
        \midrule
        \multirow{4}{*}{\Th{\vam{}-B}} & \multirow{4}{*}{318 + 38} & 38 & 2.585 & 53.6 & 1.686 & 40.9 \\
        && 77 & 2.218 & 62.4 & 1.489 & 38.7 \\
        && 116 & 2.814 & 48.4 & 2.004 & 39.6 \\
        && 139 & 1.781 & 70.0 & 1.234 & 39.7 \\
        \midrule
        \Th{\vam{}-L} & 1,200 + 150 & 139 & 2.375 & 56.80 & 1.541 & 40.2 \\
        \bottomrule
    \end{tabularx}
    \label{tab:ncap_frontal}
\end{table*}

%% file: tex/conclusion.tex
\section{Conclusion}
\label{sec:conclusion}

\vm{} and \vam{} are a significant step forward in applying large-scale unlabeled pre-training to autonomous driving, offering several exciting discoveries. 
First, the successful transfer of the pre-trained representations to driving tasks demonstrates the versatility of our approach. Complex driving behaviors are learned directly from raw video without requiring expensive semantic annotations. Particularly encouraging is \vam{}'s reduction of existing methods' collision rates by 27\% while maintaining comparable progress metrics.
Second, the performance on out-of-distribution datasets like KITTI and Cityscapes demonstrates our approach's robustness and generalization capabilities.
Third, our scaling experiments reveal a clear path forward. The empirical scaling laws we established suggest substantial headroom for improvement, notably through increased training data.

By releasing our complete codebase, training recipes, scaling laws, and model weights, we aim to accelerate progress in video-based autonomous driving. We envision several promising directions for future work:
\begin{itemize}
    \item Decoupling high-level command path from actual expert trajectory so that, in the imitation training set, the model observes the expert deviating from the high-level command path.
    \item Extending our approach to leverage multi-camera setups for enhanced scene understanding.
    \item Exploring more sophisticated action generation frameworks that maintain the benefits of our current approach while improving safety-critical behavior.
    \item Investigating larger-scale pre-training on even more diverse driving datasets.
    \item Using a better tokenizer than the current LLaMaGen-VQGAN, adapted to driving scenarios and better able to capture fine visual details (text on signs, road markings, traffic lights, etc.).
\end{itemize}

The key limitation of our work lies in the gap between \vm{}'s ability to model future states and \vam{}'s current reliance on imitation learning. While \vm{} can generate plausible future video streams, we have not yet leveraged that predictive power for planning and control. A critical missing piece is a reward model that distinguishes between favorable and critical latent states, enabling more sophisticated planning strategies beyond pure imitation. That presents an exciting opportunity to transform our reactive system into a proper `world-model'-based planning pipeline.

Additionally, while comprehensive for driving performance, our evaluation framework does not fully evaluate the depth of physical understanding learned by our video model. Future work should develop more nuanced evaluation metrics, similar to Physics-IQ~\cite{motamed2025physicsIQ}'s spatial and temporal mIoU approach to assess different aspects of physical understanding. Such metrics would provide deeper insights into what our models learn about scene dynamics, object interactions, and physical constraints.

\section*{Acknowledgements}
\label{sec:acknowledgment}
This work was partially supported by the ANR MultiTrans project (ANR-21-CE23-0032). It was initially explored using HPC resources from GENCI–CINES (Grant 2023-A0141014181), and most of its results were obtained using HPC resources from GENCI–IDRIS (Grant 2024-GC011015459). We acknowledge the EuroHPC Joint Undertaking for awarding this project access to the EuroHPC supercomputer LEONARDO, hosted by CINECA (Italy) and the LEONARDO consortium, through a EuroHPC AI and Data-Intensive Access call.

\clearpage

\section*{Detailed Contributions}
\label{sec:credits}

{\sc Project Lead} \\
\textit{\small Research direction, technical roadmap, and project coordination} \\
Florent Bartoccioni

{\sc Core Contributors} \\
\textit{\small All aspects of the codebase, experiments, and evaluations} \\
Florent Bartoccioni, Elias Ramzi

{\sc Contributors} \\
Victor Besnier --- \textit{\small Visual Tokenization codebase using pre-trained VQGAN; FID metric code}\\
Loick Chambon --- \textit{\small Data download, transfer and extraction; visualization code development}\\
Eduardo Valle --- \textit{\small OpenDV preprocessing}\\
Shashanka Venkataramanan --- \textit{\small Depth anything pseudo-GT generation}\\
Tuan-Hung Vu --- \textit{\small GPT adaptation from nanoGPT}\\
Yihong Xu --- \textit{\small nuPlan preprocessing and initial dataloader development}

{\sc Technical Report} \\
\textit{\small Manuscript preparation, design, visualizations, figures} \\
Florent Bartoccioni, Elias Ramzi, Victor Besnier, Shashanka Venkataramanan, Eloi Zablocki, Yihong Xu, Tuan-Hung Vu, Eduardo Valle

{\sc Grant Acquisitions} \\
\textit{\small Grant proposals for Adastra, EuroHPC, and Jean Zay Grand Challenges} \\
Florent Bartoccioni, Alexandre Boulch, Mickael Chen, Eduardo Valle, Spyros Gidaris, Eloi Zablocki, Matthieu Cord, Serkan Odabas, David Hurych

{\sc Advisory} \\
\textit{\small Research and organization guidance} \\
Eloi Zablocki, Alexandre Boulch, Mickael Chen

{\sc Senior Advisory} \\
\textit{\small Research and organization guidance} \\
Eduardo Valle, Andrei Bursuc, Renaud Marlet, Matthieu Cord\\